\documentclass{article}

 \usepackage[preprint]{neurips_2026}

\PassOptionsToPackage{numbers, compress}{natbib}

\usepackage[utf8]{inputenc}
\usepackage[T1]{fontenc}
\usepackage{microtype}
\usepackage{nicefrac}
\usepackage{amsmath, amssymb, amsfonts}
\usepackage{bbold}

\usepackage[dvipsnames, table]{xcolor}
\definecolor{mygray}{gray}{.9}

\usepackage{booktabs}
\usepackage{multirow}
\usepackage{array}
\usepackage{graphicx}
\usepackage{colortbl}
\usepackage{threeparttable}
\usepackage{wrapfig}
\usepackage{epsfig}
\usepackage{float}

\usepackage[skip=0pt]{caption}
\usepackage[skip=2pt, font=scriptsize, labelfont=scriptsize]{subcaption}
\usepackage[inline]{enumitem}
\usepackage{soul}
\usepackage{cancel}
\usepackage{pifont}

\usepackage{numprint}
\usepackage{siunitx}

\usepackage{url}
\usepackage[pagebackref=true, breaklinks=true, colorlinks, bookmarks=false]{hyperref}

\usepackage{comment}
\usepackage[most]{tcolorbox}
\tcbuselibrary{breakable, skins}

\usepackage{etoolbox}
\usepackage{footmisc}
\usepackage{listings}
\usepackage{pgfplots}
\newcommand{\ubold}{\fontseries{b}\selectfont}
\robustify\ubold

\newtcolorbox{neuprompt}[1]{
  colback=gray!5,
  colframe=black!60,
  boxrule=0.4pt,
  arc=2pt,
  left=6pt,
  right=6pt,
  top=4pt,
  bottom=4pt,
  fonttitle=\bfseries\small,
  fontupper=\small\ttfamily,
  title=#1,
  breakable
}

\pgfplotsset{compat=1.18}
\definecolor{llamacolor}{rgb}{0.356,0.738,0.498}
\definecolor{clipbcolor}{rgb}{0.906,0.702,0.886}
\definecolor{sfcolor}{rgb}{0.694,0.886,0.988}
\definecolor{ivcolor}{rgb}{0.961,0.871,0.702}

\title{SVAG-Bench: A Large-Scale Benchmark for Multi-Instance Spatio-temporal \\ Video Action Grounding}

\author{
  Tanveer Hannan$^{1,2}$\thanks{Equal contribution} \quad
  Shuaicong Wu$^{1}$\footnotemark[1] \quad
  Mark Weber$^{2,3}$ \quad
  Suprosanna Shit$^{4}$ \\
  \textbf{Jindong Gu}$^{5}$ \quad
  \textbf{Rajat Koner}$^6$ \quad 
  \textbf{Aljoša Ošep}$^{7}$ \quad
  \textbf{Laura Leal-Taixé}$^7$ \quad
  \textbf{Thomas Seidl}$^{1,2}$ \\
   $^1$LMU Munich \quad
   $^2$MCML \quad
   $^3$Technical University of Munich  \quad \\
   $^4$University of Zurich \quad 
   $^5$University of Oxford \quad 
   $^6$Amazon \quad
   $^7$NVIDIA\quad
   \\
  \texttt{hannan@dbs.ifi.lmu.de, shuaicong.wu@campus.lmu.de}
}

\begin{document}

\maketitle
\begin{abstract}
A truly capable AI system must do more than detect objects or recognize activities in isolation. It must form unified, grounded representations of who is acting, what they are doing, and when and where these actions unfold. These representations provide the perceptual bedrock for high-level reasoning, planning, and embodied interaction in the real world. Building such agents is central to long-horizon goals in embodied AI and robotics. Current video benchmarks evaluate fragments of these capabilities in isolation. They focus on either spatial grounding, object tracking, or temporal localization. As a result, they cannot rigorously measure progress on their joint, multi-instance integration. We introduce Spatio-temporal Video Action Grounding (SVAG), a task and benchmark that explicitly targets this unified competence by requiring models to simultaneously detect, track, and temporally localize all objects that satisfy a natural language query in complex, multi-actor scenes. To support this task, we construct SVAG-Bench. It comprises 688 videos, 19,590 verified annotations, and 903 unique action verbs drawn from crowded urban environments, wildlife, and traffic surveillance. Each video has on average 28.5 action-centric queries. This yields the densest annotation among comparable video grounding benchmarks and enables fine-grained evaluation of multi-actor disambiguation, temporal overlap, and action compositionality. Annotations are produced by a pipeline that combines expert manual labeling, GPT-3.5 paraphrase augmentation, and human verification to ensure both linguistic diversity and correctness. We further release SVAGEval, a standardized multi-referent evaluation toolkit. We also introduce SVAGFormer, a strong modular baseline architecture for SVAG. Extensive experiments across specialist and large vision–language models reveal a substantial performance gap, especially on dense, long-form videos, indicating that unified spatio-temporal action grounding is far from solved and that existing LVLMs lack the precision needed for multi-instance video understanding. By exposing this gap in a realistic, action-centric setting, SVAG-Bench offers a timely foundation for the next generation of video, robotics, and embodied AI systems whose world models must be grounded in the rich, multi-agent dynamics of real environments.
\end{abstract}

\section{Introduction}

\begin{figure}[!t]
    \centering
    \begin{subfigure}[b]{\textwidth}
        \centering
        \includegraphics[width=0.8\textwidth]{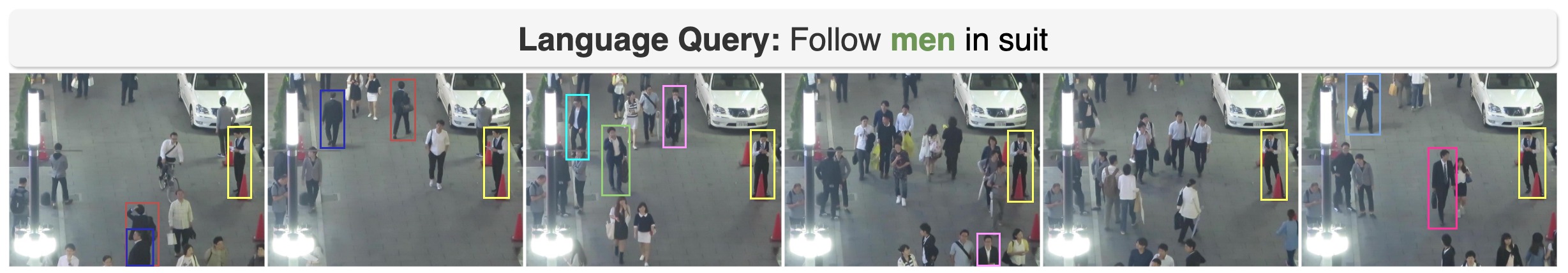}
        \caption{SVG: Spatial Video Grounding - lacks temporal grounding}
        \label{fig:intro_type}
    \end{subfigure}

    \begin{subfigure}[b]{\textwidth}
        \centering
        \includegraphics[width=0.8\textwidth]{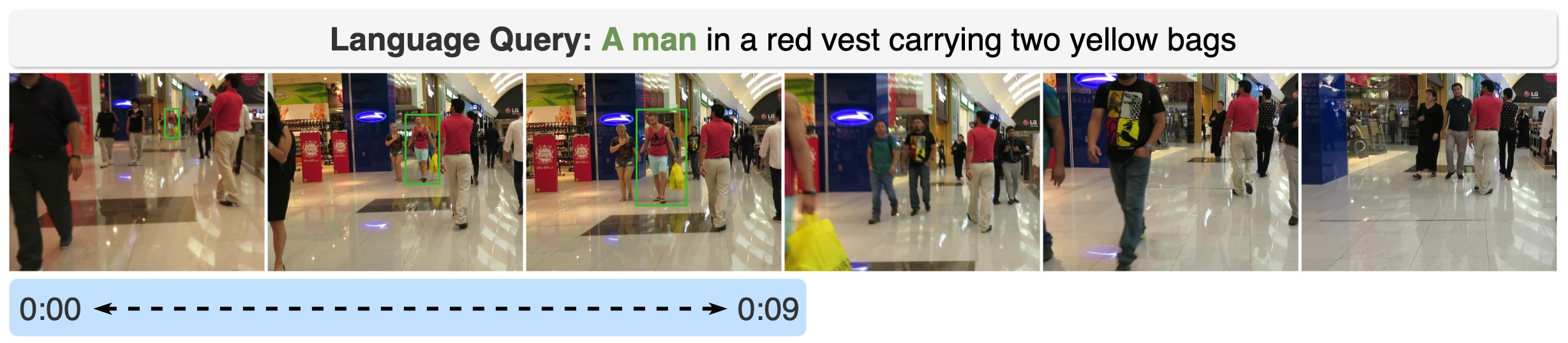}
        \caption{STVG: Spatio-Temporal Video Grounding - lacks multi-instance grounding}
        \label{fig:intro_stvg}
    \end{subfigure}

    \begin{subfigure}[b]{\textwidth}
        \centering
        \includegraphics[width=0.8\textwidth]{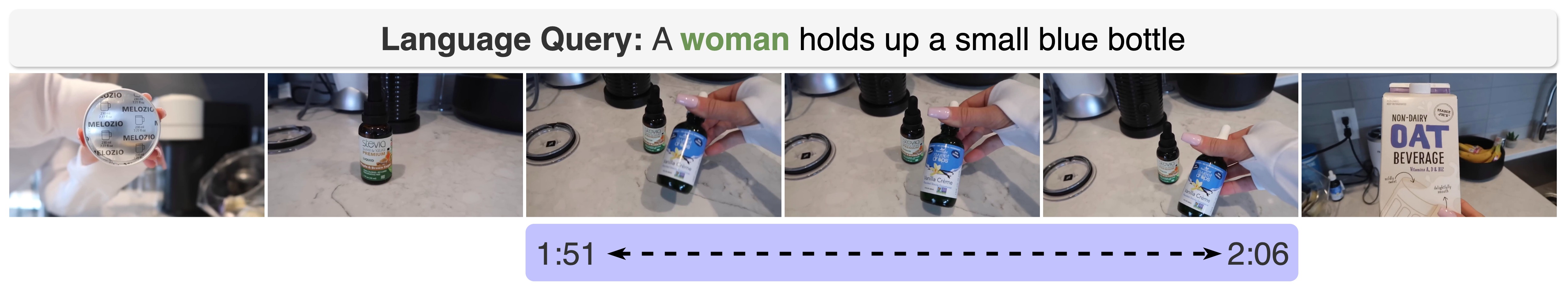}
        \caption{VTG: Video Temporal Grounding - lacks spatial localization and multi-instance grounding}
        \label{fig:intro_vtg}
    \end{subfigure}

    \begin{subfigure}[b]{\textwidth}
        \centering
        \includegraphics[width=0.8\textwidth]{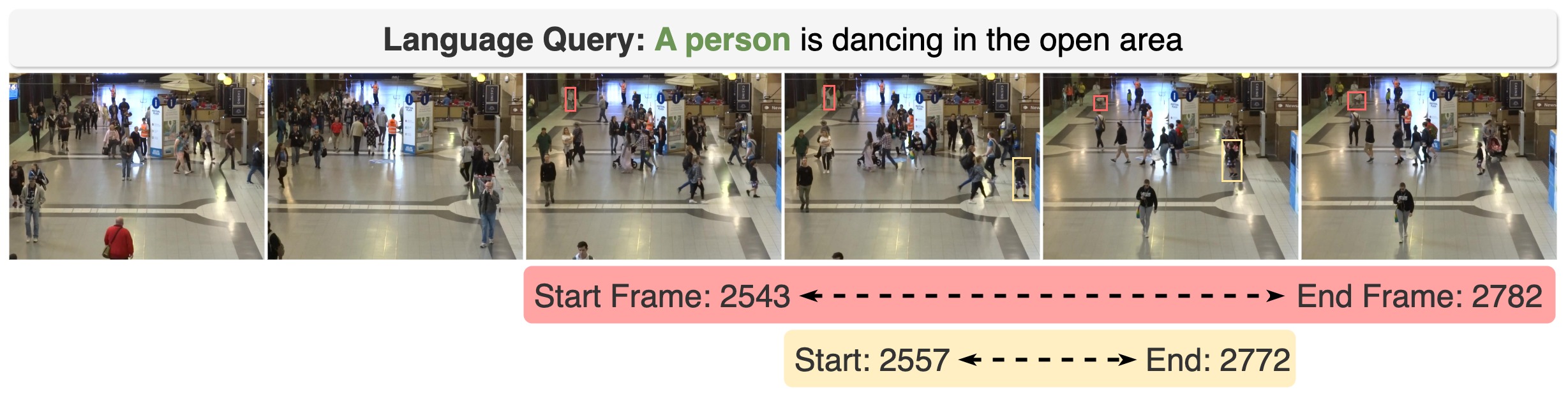}
        \caption{\textbf{Ours}: Spatio-temporal Video Action Grounding}
        \label{fig:intro_svag}
    \end{subfigure}

    \caption{
    Comparison of existing video grounding paradigms with our proposed \textbf{Spatio-temporal Video Action Grounding (SVAG)} task.
    (a) \textbf{SVG}: Spatial Video Grounding focuses only on spatial localization and lacks temporal reasoning.
    (b) \textbf{STVG}: Spatio-Temporal Video Grounding jointly localizes objects over time but cannot handle multiple interacting instances.
    (c) \textbf{VTG}: Video Temporal Grounding identifies temporal segments but misses spatial localization.
    (d) \textbf{Ours (SVAG)}: Unifies temporal and spatial grounding to detect and track multiple referent objects performing the queried action across time.
    }
    \label{fig:intro3tasks}
\end{figure}

The ability to jointly perceive \emph{who} is acting, \emph{what} they are doing, and \emph{when and where} that action unfolds is a prerequisite for AI systems that must operate in the physical world. Autonomous robots navigating crowded spaces, assistive agents monitoring patient activity, and multi-camera surveillance systems all depend on this joint capability~\cite{lee2024enhancing, liu2024end, shi2023temporal}. Yet today's video understanding models cannot reliably perform it, and more critically, the community has no shared standard to measure whether progress is being made.

Progress in video-language understanding has so far been organized around three complementary but isolated sub-tasks, as illustrated in Fig.~\ref{fig:intro3tasks}. Spatial Video Grounding (SVG)~\cite{fan2019lasot, seo2020urvos} anchors objects in space by appearance but is oblivious to when or why they act. Video Temporal Grounding (VTG)~\cite{lei2021detecting, gao2017tall, regneri2013grounding} identifies moments in time but discards the spatial extent of actors entirely. Spatio-Temporal Video Grounding (STVG)~\cite{zhang2020does, tang2021human} begins to unify these views yet remains confined to a single referent per query in short, controlled clips. Because each paradigm is evaluated in isolation, models can excel on all three benchmarks and still fail on the scenarios that matter most in deployment. Identifying when a worker skips a safety step on a busy assembly line while colleagues perform the correct motion around them, or when a single resident in a care facility pockets their medication while others at the same table swallow theirs, these one-among-many, temporally overlapping situations define real-world embodied perception, and no existing benchmark measures them.
% the moment a real scene presents two people dancing simultaneously, or a dog chasing a chicken while a second dog watches.

This fragmentation is not a minor inconvenience; it is a structural barrier on the path toward embodied intelligence. Embodied agents and robotic perception systems require \emph{compositional action understanding}: the capacity to associate multiple co-occurring actors with their respective actions and to track each one across time, all from a single natural language instruction~\cite{gu2024context, zhang2020does, tang2021human}. Without a benchmark that probes this capability end-to-end, the field cannot determine whether a new model has genuinely acquired it or has merely learned to exploit the statistical regularities of narrow, single-referent datasets. The absence of a suitable evaluation standard is therefore not just a measurement problem; it actively conceals a capability gap that matters deeply for physical AI systems.

To close this gap, we introduce \textbf{Spatio-temporal Video Action Grounding (SVAG)}, a new task and benchmark designed to operationalize holistic, action-centric video understanding. Unlike prior settings, SVAG requires a model to (1) detect \emph{all} objects performing the queried action, (2) track their spatial positions across time, and (3) precisely localize the temporal intervals in which those actions occur. Given the query ``A person is dancing in the open area,'' multiple individuals may satisfy the description simultaneously, and the model must recover the spatial and temporal extent of each one, as shown in Fig.~\ref{fig:intro_svag}. This formulation reflects the compositional and multi-agent structure of real-world scenes in a way that no existing paradigm captures.

To operationalize SVAG, we introduce \textbf{SVAG-Bench}, a large-scale benchmark built exclusively around \emph{action-centric} queries. Where prior datasets ask ``which object matches this description?'', SVAG-Bench asks ``which objects are doing this, where and for how long?'' This shift in question type drives a fundamentally different reasoning demand: models must understand motion patterns, temporal evolution, and inter-object interactions rather than static appearance. The dataset contains 9,781 manually annotated video-query pairs covering 480 distinct action verbs; query augmentation via GPT-3.5~\cite{openai2023} expands this to 19,590 records encompassing 903 unique verbs, ranging from atomic actions (\textit{The person walks inside the boat}) to complex multi-actor interactions (\textit{The horse which runs past other horses in a race}). A statistical comparison with existing datasets is presented in Table~\ref{tab:dataset_comparison}. With an average of 28.5 queries per video, roughly $7\times$ the average density of prior benchmarks, SVAG-Bench enables fine-grained evaluation of temporal overlap, multi-actor disambiguation, and action compositionality that existing resources cannot support.

\begin{table}[!t]
\centering
\renewcommand{\arraystretch}{1.1}
\setlength{\tabcolsep}{6pt}
\scalebox{0.82}{
\begin{tabular}{lcccccc}
\specialrule{1.3pt}{0pt}{0pt}
\textbf{Dataset} & \textbf{Videos} & \textbf{Queries} & \textbf{Tracks} & \textbf{Queries / Video} & \textbf{Tracks / Video} & \textbf{Distinct Verbs} \\
\specialrule{1pt}{0pt}{0pt}
Refer-Youtube-VOS~\cite{seo2020urvos} & 3,978 & 14,952 & 7,451 & 3.76 & 1.87 & 876 \\
GroOT~\cite{nguyen2023type} & 1,515 & 3,567 & 13,294 & 2.35 & 8.77 & 197 \\
VidSTG~\cite{zhang2020does} & \textbf{6,924} & \textbf{44,808} & \textbf{35,044} & 6.47 & 5.06 & 246 \\
HC-STVG~\cite{tang2021human} & 5,660 & 5,660 & 5,660 & 1.00 & 1.00 & 515 \\
\hline
\textbf{SVAG-Bench (Ours)} & 688 & 19,590 & 9,781 & \textbf{28.47} & \textbf{14.22} & \textbf{903} \\
\specialrule{1.3pt}{0pt}{0pt}
\end{tabular}}
\vspace{2mm}
\caption{
\textbf{Comparison of video grounding datasets.}
Refer-Youtube-VOS and GroOT belong to the SVG domain, while VidSTG and HC-STVG are STVG datasets.
Although VidSTG has the largest overall scale, \textbf{SVAG-Bench} achieves the highest \textit{annotation density} (queries and tracks per video) and the broadest \textit{action diversity} (distinct verbs), making it particularly suited for fine-grained, multi-object spatio-temporal grounding.
}
\label{tab:dataset_comparison}
\vspace{-3mm}
\end{table}

Datasets such as Refer-Youtube-VOS and GroOT belong to the \textit{spatial grounding} domain, while VidSTG and HC-STVG address \textit{spatio-temporal grounding} with a single referent per query. On average, these benchmarks contain only {3.8 queries per video}, indicating sparse supervision and limited linguistic diversity. In contrast, SVAG-Bench provides {28.5 queries per video}, offering significantly denser annotations across queries, tracks, and actions. This high annotation density enables fine-grained evaluation of temporal overlap, multi-actor disambiguation, and action compositionality. Further details of the annotation pipeline, statistics, and taxonomy are discussed in Section~\ref{sec:dataset}.

We believe that benchmarks which demand joint spatio-temporal action understanding are not merely useful additions to the video understanding community; they are necessary infrastructure for embodied AI research~\cite{gu2024context, khoreva2019video, nagaraja2016modeling, wu2023referring}. A robot that can parse ``stop the person running toward the train'' must solve exactly the problem SVAG formalizes: it must ground an action description to a specific actor, locate that actor spatially, and determine when the described event is occurring. As foundation models are increasingly integrated into robotic perception pipelines, the demand for standardized evaluations of this joint capability will only grow. SVAG-Bench is designed to serve as that standard. We have open-sourced the dataset, model, and evaluation toolkit for further improvement of Embodied AI.

\textbf{Summary of Contributions.}

\begin{enumerate}
    \item \textbf{New Task.} Spatio-temporal Video Action Grounding (SVAG):
    a unified formulation that jointly requires multi-object tracking, action understanding, and temporal localization across multiple referents.

    \item \textbf{New Dataset.} SVAG-Bench:
    a large-scale, dense, action-centric benchmark with 19,590 annotated records, 903 unique verbs, and 28.5 queries per video, designed to expose the compositional reasoning gap in current video models.

    \item \textbf{New Baseline.} SVAGFormer:
    a transformer framework that jointly integrates spatial localization and temporal grounding for the SVAG task.

    \item \textbf{New Evaluation Protocol.} SVAGEval:
    a standardized toolkit for benchmarking multi-referent spatio-temporal grounding, providing a reproducible platform for future studies. 
\end{enumerate}
\section{Related Work}
\vspace{-1em}
\noindent \textbf{Spatial Video Grounding (SVG)} extends multi-object tracking~\cite{hannan2023gratt,
koner2023instanceformer, hannan2022box} to language-guided localization, where one
or more objects must be identified in a video based on natural language descriptions,
typically via bounding boxes or pixel-level masks.
Existing SVG benchmarks~\cite{fan2019lasot, seo2020urvos, xiao2020visual} rely
predominantly on static visual attributes such as category, color, or position,
allowing the target to be uniquely identified from the very first frame
(e.g., ``track the man in a suit'') without any temporal reasoning.
As a result, conventional detectors and trackers often suffice, and
the dynamic behavior of objects over time is largely ignored. Recent efforts such as GroOT~\cite{nguyen2023type} extend SVG to include
action-related queries, but their scope remains narrow, covering few verbs and
simple single-object interactions.
Subsets of TAO~\cite{dave2020tao} introduce multi-object tracking but focus on
appearance-based distinctions rather than motion or intent.
Models such as Referring Multi-Object Tracking (RMOT)~\cite{wu2023referring,
zhang2024bootstrapping} generalize SVG to multiple referents but remain restricted
to short, domain-specific videos such as traffic and pedestrian scenes.
SVAG-Bench explicitly targets \textit{action-driven semantics}, where multiple
visually similar entities must be distinguished by their dynamic behavior across
time, a demand that appearance-based SVG formulations are structurally unable
to satisfy.

\noindent \textbf{Video Temporal Grounding (VTG)} focuses on localizing temporal segments in
untrimmed videos that correspond to natural language queries~\cite{hannan2024rgnet,
hannan2025revisionllm, qu2024chatvtg, lin2023univtg, chen2021end, xu2022point}.
Benchmarks such as QVHighlights~\cite{lei2021detecting},
Charades-STA~\cite{gao2017tall}, and TACoS~\cite{regneri2013grounding} frame this
as identifying the start and end times of queried events through Moment Retrieval
or Grounded question answering~\cite{hannan2025docslm}.
While these datasets have driven substantial advances in temporal reasoning, they
annotate only a single event instance per query and discard the spatial extent of
actors entirely, making it impossible to evaluate whether a model knows
\textit{who} is performing the action, only \textit{when} it occurs. SVAG extends this formulation to the object level, requiring models to jointly
perform detection, tracking, and temporal localization for all entities satisfying
the query.
A single query may correspond to multiple distinct actors or non-contiguous time
intervals, reflecting real-world scenes where several people perform the same action
at different moments.
This multi-instance, temporally compositional structure is precisely the capability
that VTG benchmarks cannot probe, and that embodied perception systems require.

\noindent \textbf{Spatio-Temporal Video Grounding (STVG)} generalizes SVG by jointly predicting spatial
trajectories and temporal segments for the object referred to in a query~\cite{
zhang2020does, tang2021human, wang2025spacevllm, luo2025spatial, lei2019tvqa+}.
This is the closest prior paradigm to SVAG, yet existing STVG datasets are confined
to a single referent per query in short, visually simple clips.
Queries are dominated by static appearance cues (e.g., ``the man in a red shirt'')
or coarse action labels (e.g., ``a person jumping''), which limits their ability to
evaluate fine-grained, compositional, or multi-actor behaviors.
STVGFormer~\cite{lin2022stvgformer} exemplifies this trend, relying primarily on
visual appearance for spatial grounding and using only coarse human-action cues
for temporal reasoning.
VidSTG~\cite{zhang2020does}, the largest STVG benchmark, derives its sentences from
fixed subject-predicate-object triplets in VidOR~\cite{shang2019annotating},
which structurally constrains both linguistic and action diversity. SVAG-Bench moves beyond these constraints through densely annotated, action-centric
queries that span longer temporal horizons and visually complex scenes.
It explicitly requires spatial and temporal grounding of \textit{all} objects
satisfying a query, enabling evaluation of multi-instance, multi-action understanding
that STVG benchmarks, by construction, cannot provide.
This is not an incremental extension; it is the missing evaluation layer between
current video grounding research and the joint perceptual reasoning that
robotic and embodied AI systems demand.% 

\section{Dataset Overview}
\label{sec:dataset}

\subsection{Data Collection and Annotation}

To support the proposed SVAG task, we construct SVAG-Bench, a comprehensive benchmark designed to cover a broad range of scenes, object categories, and action types.  
Videos are curated from multiple real-world domains, including crowded urban environments, traffic surveillance, wildlife monitoring, and natural ecosystems.  
Our goal in building SVAG-Bench is two-fold: to ensure \textit{completeness}, by including diverse interaction patterns and environments, and to ensure \textit{discrimination}, by featuring multiple visually similar instances of the same category engaged in distinct actions.

The dataset sources videos from established multi-object tracking benchmarks, e.g.,MOT17~\cite{dendorfer2021motchallenge}, MOT20~\cite{dendorfer2020mot20}, and OVIS~\cite{qi2022occluded}, where we select sequences in which objects of similar appearance perform different actions.  
Human annotators manually label all visible objects with concise, action-centric natural language descriptions.  
To enhance linguistic richness and generalization, we further expand these annotations using paraphrases generated by GPT-3.5~\cite{openai2023}, followed by human verification to ensure correctness and naturalness.

In total, \textbf{SVAG-Bench} comprises approximately \textbf{19,590 action-based annotations} across \textbf{688 videos}, providing fine-grained ground truth for both spatial and temporal grounding.  
This combination of dense action coverage and linguistic variation enables robust evaluation of multi-object, multi-action understanding. Detailed annotation procedures, quality assurance protocols, and representative examples are provided in Appendix~\ref{appendix:annotation}, and ~\ref{appendix:preprocess}.

\begin{figure}[!t]
  \centering
  % 左图
  \begin{subfigure}[b]{0.47\textwidth}
    \centering
    \includegraphics[width=\linewidth]{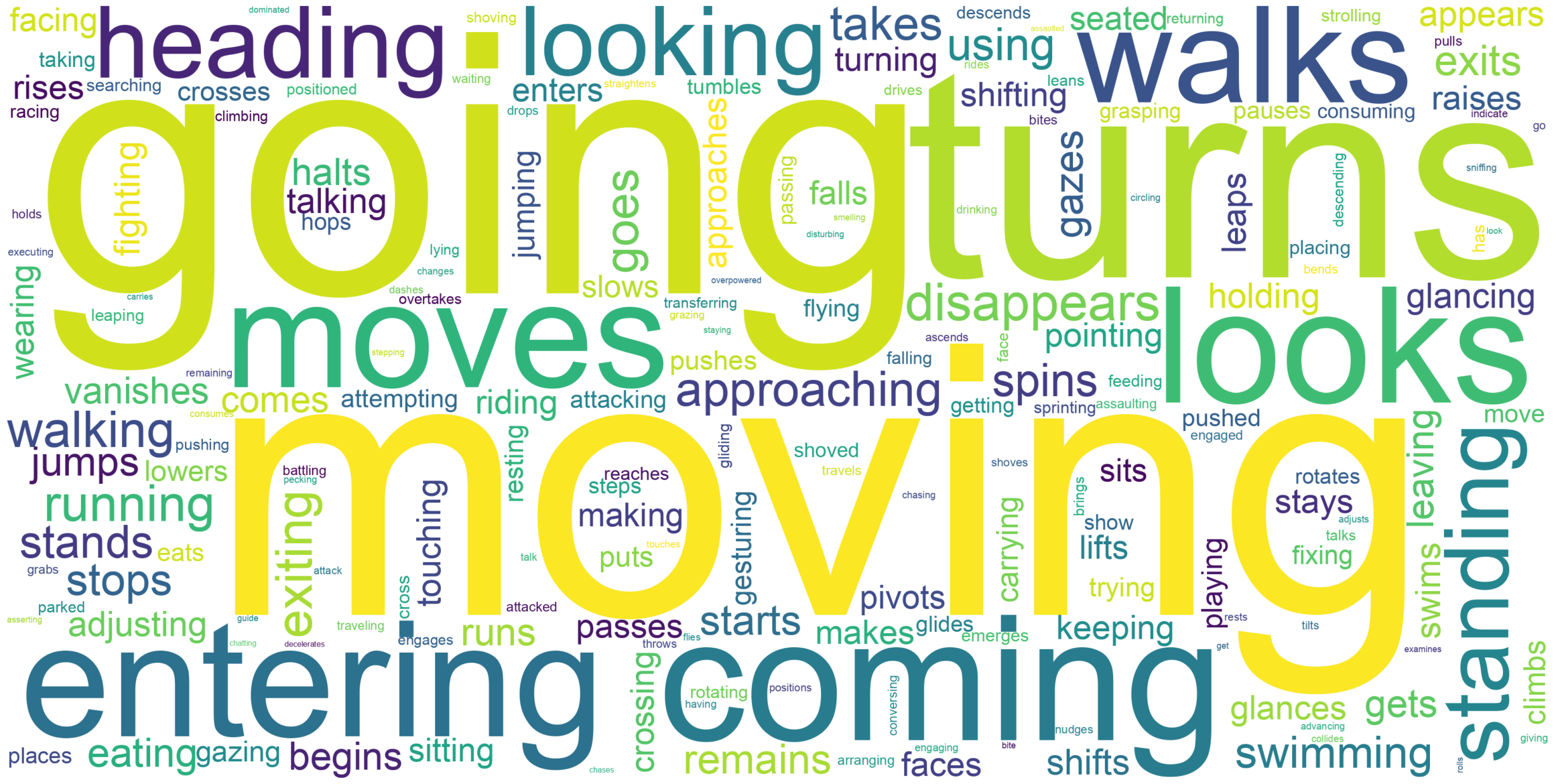}
    \caption{Word cloud of the verbs}
    \label{fig:wordcloud}
  \end{subfigure}\hfill
  % 右图
  \begin{subfigure}[b]{0.52\textwidth}
    \centering
    \includegraphics[width=\linewidth]{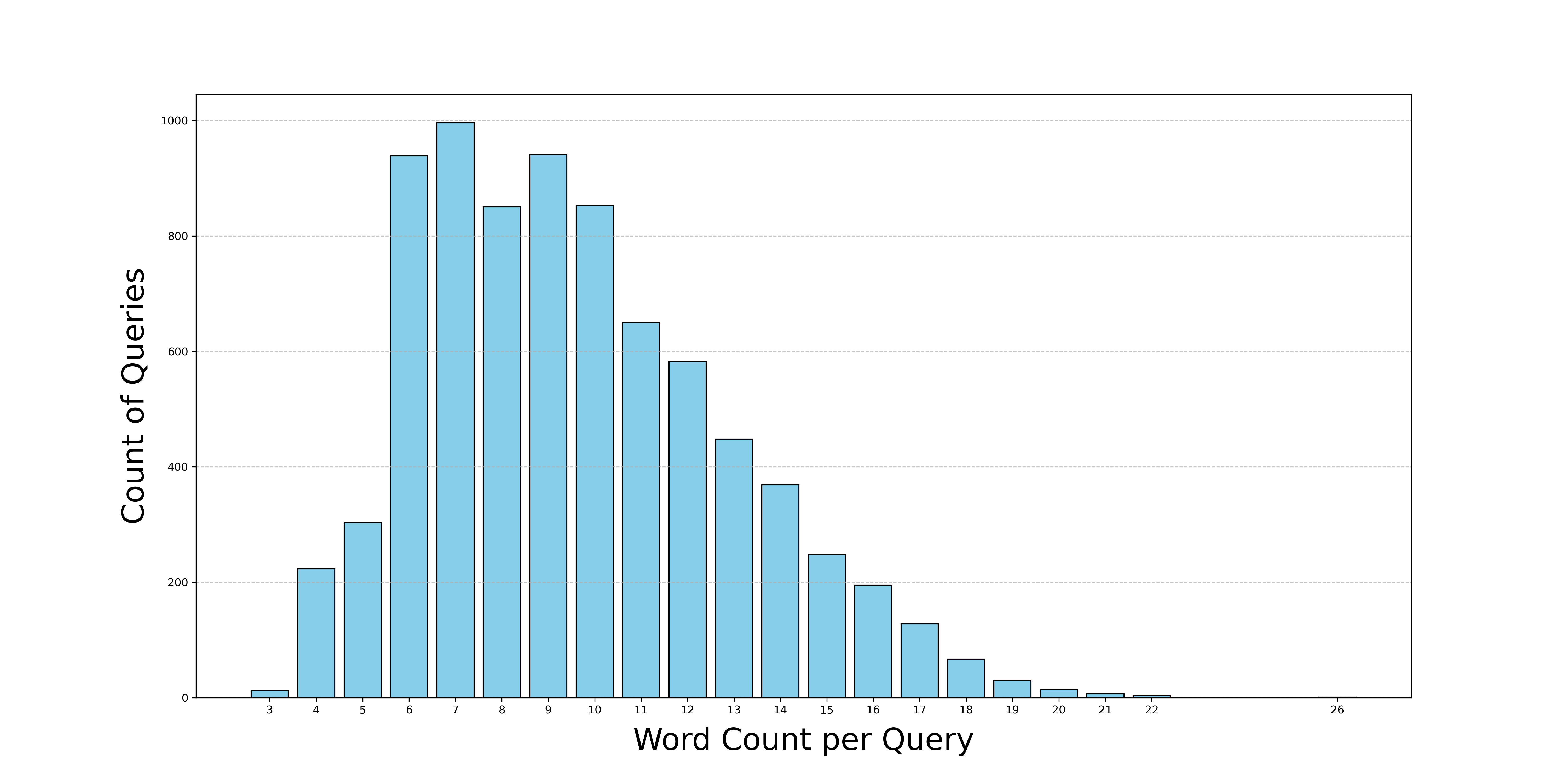}
    \caption{Distribution of query word counts}
    \label{fig:query_word-counts}
  \end{subfigure}

  \caption{\textbf{Statistics of SVAG-Bench}. The majority of queries fall within the range of 6 to 10 words.}
  \label{fig:statistics}
\end{figure}

\subsection{Dataset Statistics}
To assess the richness and complexity of SVAG-Bench, we conduct detailed statistical analyses and compare it with representative benchmarks from the \textit{Spatial Video Grounding (SVG)} and \textit{Spatio-Temporal Video Grounding (STVG)} domains.  
A summary of this comparison is presented in Table~\ref{tab:dataset_comparison}.  
Overall, SVAG-Bench exhibits the highest values across key indicators such as \textit{queries per video (28.47)}, \textit{tracks per video (14.22)}, and \textit{distinct verbs (903)}, highlighting its superior annotation density, object diversity, and action coverage.

Unlike prior datasets that emphasize total size (i.e., total number of videos or queries), \textbf{per-video annotation density} serves as a more meaningful measure of complexity and reasoning difficulty.  
High-density videos introduce frequent interactions between multiple entities and actions, requiring models to perform precise spatio-temporal reasoning under dense and overlapping conditions.  
This property makes SVAG-Bench particularly suitable for evaluating fine-grained video-language understanding beyond appearance-based or single-actor scenarios.

In summary, SVAG-Bench achieves a unique balance between scale and annotation depth: it is compact enough for efficient experimentation yet dense and diverse enough to challenge current vision-language models with realistic, multi-object, action-centric reasoning tasks.

\textbf{Distinct Verbs (Action).} Since our task focuses on object grounding based on actions, the diversity of verb usage in natural language queries is a key metric in evaluating annotation quality. To quantify verb diversity consistently across datasets, we adopt a unified methodology:

\begin{enumerate}
    \item We use the \href{https://spacy.io/}{spaCy} library to tokenize and parse all natural language queries, extracting all tokens with a POS tag of VERB.
    \item For sentences containing multiple consecutive verbs (e.g., ``a person stops walking''), we retain only the main action verb and remove the preceding verbs such as ``stops'', ``starts''.
    \item Verb diversity is computed as the number of unique verb lemmas across all queries.
\end{enumerate}

A word cloud of all annotated verbs is shown in Fig.~\ref{fig:wordcloud} to visualize the linguistic diversity. The cloud includes verbs in different tenses (e.g., “moving”, “turns”). This level of lexical variability enhances the usefulness of this dataset for training and evaluating language-based video understanding models. As can be seen in Table~\ref{tab:dataset_comparison}, SVAG's 903 actions is more than the other three datasets (GroOT: 197, VidSTG: 246, and HC-STVG: 515), and slightly higher than Ref-Youtube-VOS (876).

\textbf{Language Query Length.} The complexity of natural language queries is another important factor in assessing annotation diversity. We analyze the distribution of query lengths in terms of word count, as depicted in Fig.~\ref{fig:query_word-counts}. The results indicate that the majority of queries fall within the range of 6 to 10 words, striking a balance between conciseness and descriptive richness. The average length of all queries is about 9.58 words. This allows the queries to be both informative and tractable for grounding models. The properties of the subsets of SVAG-Bench and a detailed statistical breakdown across other aspects of the dataset are provided in Appendix~\ref{appendix:statistics}.

\section{SVAGFormer}
\begin{figure*}[!t]
  \begin{center}
    \includegraphics[width=0.82\linewidth]{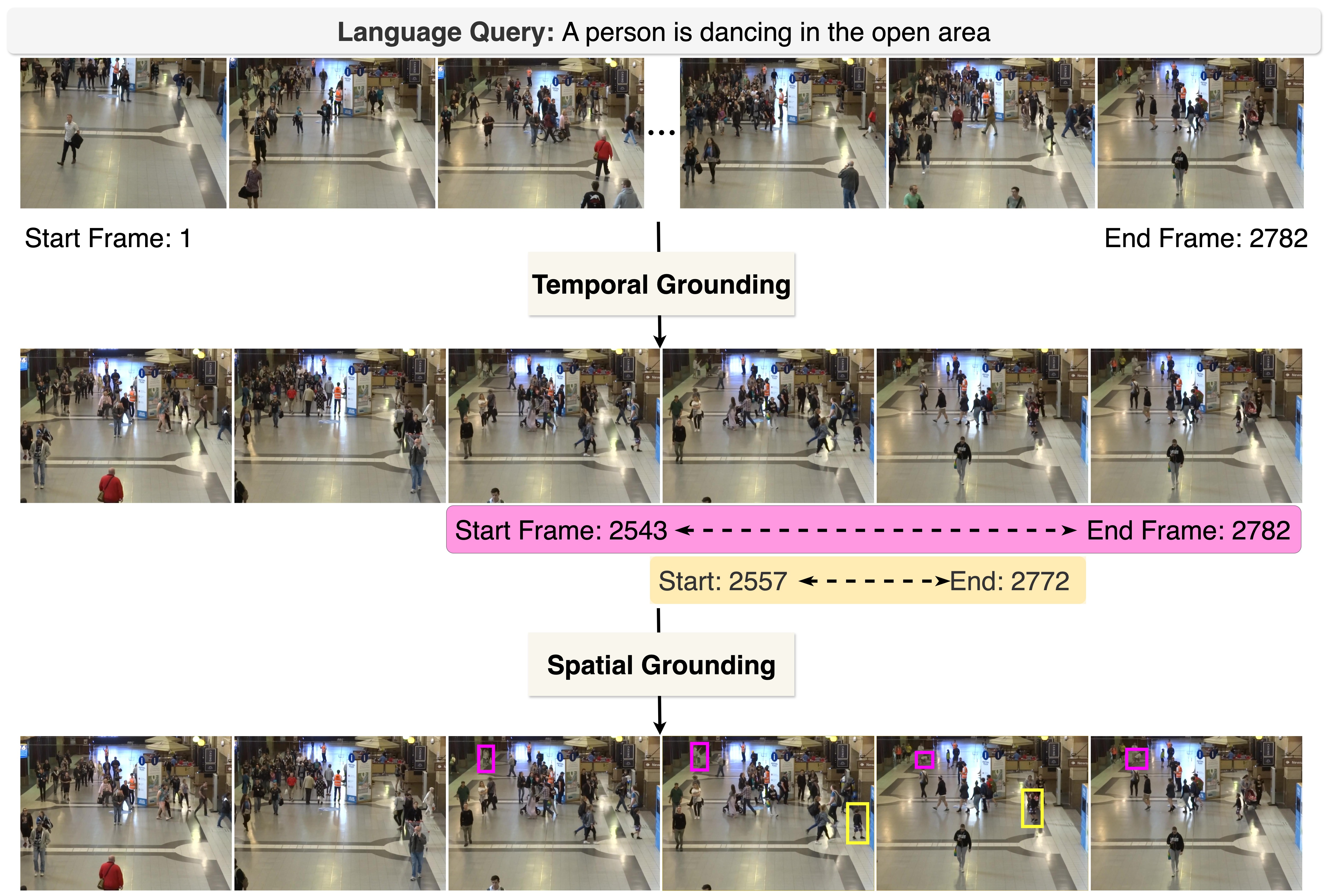}
  \end{center}
  \caption{Overview of the \textbf{SVAGFormer} pipeline. Given a natural
  language query (e.g., ``A person is dancing in the open area''), temporal
  grounding first narrows the full video (frames 1--2782) to two temporal
  candidates (2543--2782) and (2557--2772).
  Spatial grounding then operates exclusively within this window, returning
  bounding box tracks for all actors satisfying the query.
  \label{fig:method}}
\end{figure*}

We propose \textbf{SVAGFormer}, a Transformer based baseline for the SVAG task built
around a \textit{temporal-first gating} strategy: temporal grounding is
performed first to identify the action interval, and spatial grounding is
then constrained to operate exclusively within that window.
This ordering is not arbitrary. A tracker applied across the full video
accumulates detections from frames where the queried action is absent,
producing outputs that are incorrect by definition under the SVAG evaluation
protocol. Gating the spatial module with a predicted temporal boundary
enforces consistency at the architecture level rather than as a
post-processing step. For temporal grounding, we adopt FlashVTG~\cite{Cao_2025_WACV}, which
produces a coarse-to-fine action segment via temporal feature layering and
adaptive score refinement. For spatial grounding, we employ
TempRMOT~\cite{zhang2024bootstrapping}, which tracks multiple referents described
by action-oriented queries using query memory for temporal consistency.
The combined output, spatial trajectories paired with temporal
intervals for all entities satisfying the query constitute a strong
spatiotemporal baseline for SVAG. An illustration of the pipeline can be found in Fig.~\ref{fig:method}.

\section{Experiments}
\subsection{Evaluation Metrics}
To comprehensively evaluate the performance of our proposed SVAG task, we adopt a set of well-established metrics tailored to the two core subtasks: spatial grounding and temporal grounding. Each subtask requires different aspects of performance to be measured, and thus, the metrics are employed accordingly. We utilize Higher Order Tracking Accuracy (HOTA)~\cite{luiten2021hota} to evaluate spatial grounding, i.e, detection in one frame and their temporal association across frames. In referring multi-object tracking (RMOT)~\cite{wu2023referring}, predicted tracks corresponding to visible objects not referenced by any query are treated as false positives, ensuring evaluation focuses only on objects relevant to the natural language query. The overall HOTA is computed by averaging per-query HOTA across all sentence queries in the dataset~\cite{wu2023referring}.
% \mar{HOTA is for tracking not spatial tasks?} \wu{...} 
For the temporal grounding task, we use Recall at 1, 5, and 10 (R1@X, R5@X, R10@X), mean Average Precision (mAP), and mean Intersection over Union (mIoU) as evaluation metrics, where X denotes the IoU threshold, specifically set to 0.1, 0.3, and 0.5.
We define our evaluation metric based on these popular metrics for spatial~\cite{wu2023referring} and temporal grounding~\cite{Cao_2025_WACV, jiang2024prior, lei2021detecting} from previous works. 
% The detailed definition and formula are provided in Appendix~\ref{appendix:metrics}. 

\subsection{SVAGEval}

We introduce SVAGEval to formalize the evaluation for the SVAG task. This official evaluation codebase also served as the benchmark for an ICCV 2025 workshop competition. Unlike existing temporal grounding protocols, SVAGEval supports multiple referents under spatiotemporal constraints. Specifically, spatial and temporal grounding are evaluated separately, with identity mapping strategies ensuring consistent alignment across the two dimensions. The final leaderboard score is computed as the arithmetic mean over OVIS, MOT17, and MOT20. Below, we describe core implementation details of our evaluation pipeline.

\begin{figure}[!h]
  \begin{center}
    \includegraphics[width=1\linewidth]{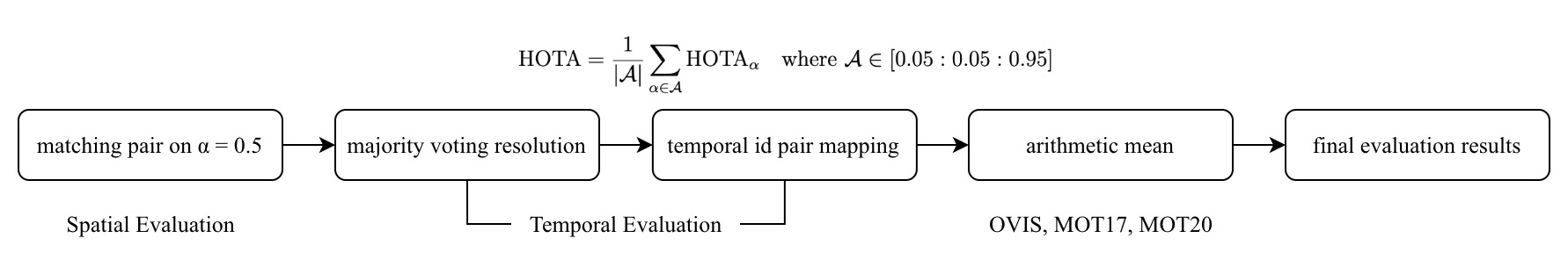}
  \end{center}
  \caption{Flowchart for processing evaluation. Spatial and temporal evaluations are conducted separately on the OVIS, MOT17, and MOT20. 
  % \mar{The dataset don't do this} 
  The results are averaged and combined to form the final result. Threshold $\alpha$ controls the relative importance of detection and association accuracy in HOTA.
  % \mar{maybe explain what alpha is?} \wu{modified}
    \label{fig:svag_svageval}
  }
\end{figure}

\textbf{Localisation Thresholds.} The HOTA \cite{luiten2021hota} score is calculated by averaging over a range of threshold values $\alpha$ that define the matching criteria between predicted and ground truth instances. Its core calculation formula is as follows:

\begin{equation}
\text{HOTA} = \frac{1}{|\mathcal{A}|} \sum_{\alpha \in \mathcal{A}} \text{HOTA}_\alpha
% \quad \text{where } \mathcal{A} \in [0.05:0.05:0.95] 
\end{equation}
where $\mathcal{A}$ denotes a set of thresholds ranging from 0.05 to 0.95 in increments of 0.05. In our implementation, we further select the matching result at $\alpha = 0.5$ as the basis for ID mapping, as it provides a balanced trade-off between strictness and flexibility in evaluating spatial matches.

\textbf{ID Mapping for Temporal Evaluation.} Since spatial and temporal grounding are evaluated separately, it is necessary to establish a reliable mapping between predicted and ground truth identities (track\_ids) before temporal evaluation. Our strategy proceeds as follows:

\begin{enumerate}
    \item Spatial ID Matching: Using the HOTA matching result at $\alpha = 0.5$, we determine a one-to-one mapping between each ground truth track\_id and its most likely predicted counterpart across frames.
    \item Majority Voting Resolution: Due to inaccurate predictions in spatial grounding, a single ground truth track\_id might be associated with multiple predicted track\_ids over time. To address this issue, we perform a majority voting scheme. For each ground truth ID, the number of frames in which each predicted ID appears is taken as the frequency, and the prediction ID with the highest frequency is selected to match the ground truth ID.
    \item Temporal Pair Construction: Using this final track\_id mapping, we find temporal prediction and ground truth pairs for each referent. These pairs are then passed into the temporal grounding evaluation module.
\end{enumerate}

Since the evaluation metrics for OVIS, MOT17, and MOT20 are calculated independently, we adopt an arithmetic mean of the scores from these datasets to produce the final metric displayed on the competition website. See Fig.~\ref{fig:svag_svageval} for the process.

This design ensures consistent identity alignment across spatial and temporal dimensions, providing a fair basis for evaluating SVAG models on complex multi-object and multi-referent scenarios.

\subsection{Implementation Details}
We follow the official training configuration of
TempRMOT~\cite{zhang2024bootstrapping} on Refer-KITTI-V2, with a
memory length of 5. We use the Adam optimizer with an initial learning
rate of $1\times10^{-5}$, decayed by a factor of 10 after epoch 40,
and train for 60 epochs on 4 GPUs. We follow the official configuration of FlashVTG~\cite{Cao_2025_WACV},
converting all data into QVHighlights format. Video and text features
are extracted via InternVideo2~\cite{wang2024internvideo2} and
LLaMA~\cite{touvron2023llama}, respectively. All feature dimensions
are set to 256, the fusion module uses 8 attention heads with $K{=}4$,
and temporal feature layering uses 5 layers. We apply AdamW as the
optimizer with an NMS threshold of 0.7. The maximum visual length is
adjusted per dataset, and models are trained independently on OVIS,
MOT17, and MOT20. For LVLMs, input frames are either resized or sampled at a fixed
count. \textit{Resize} denotes aspect-ratio-preserving resizing with
the longer side capped at 448 pixels and the shorter side ensured to
be at least 252 pixels. Fixed values (e.g., 32, 96, 128, 256) denote the
number of uniformly sampled frames. Qwen2.5-VL refers to
Qwen2.5-VL-7B-Instruct and VTimeLLM refers to VTimeLLM-7B. EgoMask uses VideoLISA~\cite{bai2024one}, an open-source VideoLLM. EgoMask uses a stride of 5 on both MOT17 and MOT20. Sa2VA uses a stride of 2 on MOT17-03 and MOT20-03, and a stride of 3 on MOT20-05.

\begin{table*}[!t]
    \centering
    \renewcommand{\arraystretch}{1.1}
    \resizebox{\textwidth}{!}{
    \begin{tabular}{ll ccc ccc ccc c}
        \specialrule{1.5pt}{0pt}{0pt}
        & & \multicolumn{3}{c}{\textbf{OVIS}} 
          & \multicolumn{3}{c}{\textbf{MOT17}} 
          & \multicolumn{3}{c}{\textbf{MOT20}} 
          & \\
        \cmidrule(lr){3-5} \cmidrule(lr){6-8} \cmidrule(lr){9-11}
        \multirow{-2}{*}{Category} & \multirow{-2}{*}{Methods} 
          & HOTA$\uparrow$ & R1@.5$\uparrow$ & mIoU$\uparrow$
          & HOTA$\uparrow$ & R1@.5$\uparrow$ & mIoU$\uparrow$
          & HOTA$\uparrow$ & R1@.5$\uparrow$ & mIoU$\uparrow$
          & \multirow{-2}{*}{m-HIoU$\uparrow$} \\
        \hline\hline

        \multirow{6}{*}{Temporal}
        & LD-DETR~\cite{zhao2025ld}
          & --    & \textbf{35.25} & \textbf{40.83}
          & --    & 4.21           & 8.90
          & --    & 0.69           & 0.55           & -- \\
        & $\rm R^2$-Tuning~\cite{liu2024r}
          & --    & 34.92          & 38.50
          & --    & \underline{8.42}  & \underline{11.60}
          & --    & \textbf{6.02}  & \underline{7.33}           & -- \\
        & VTimeLLM~\cite{huang2024vtimellm}
          & --    & 20.84          & 27.73
          & --    & 3.21           & 9.31
          & --    & 0.93           & 6.17           & -- \\
        & d2vlm~\cite{zeng2025factorized}
          & --    & 21.62          & 29.30
          & --    & 4.41           & 7.67
          & --    & 0.00           & 0.00           & -- \\
        & TimeSuite~\cite{zeng2024timesuite}
          & --    & 12.31          & 18.21
          & --    & 1.00           & 1.60
          & --    & 0.00           & 0.15           & -- \\

        \hline

        \multirow{3}{*}{Spatial}
        & Sa2VA~\cite{yuan2025sa2va}
          & 7.76           & -- & --
          & 0.45           & -- & --
          & 0.15           & -- & --  & -- \\
        & DKGTrack~\cite{li2025language}
          & \underline{17.85} & -- & --
          & \textbf{1.59}  & -- & --
          & 0.06           & -- & --  & -- \\
        & TransRMOT~\cite{wu2023referring}
          & 2.83           & -- & --
          & 0.13           & -- & --
          & \textbf{0.43}  & -- & --  & -- \\
        \hline

        \multirow{3}{*}{\shortstack[l]{Proprietary\\Spatiotemporal}}
        & GPT-4.1 mini~\cite{openai_gpt41mini_2025}
          & 4.75  & 27.30 & 30.80
          & 0.01  & 6.58  & 9.36
          & 0.00  & 0.74  & 2.36   & 7.88 \\
        & Claude-Opus-4.6~\cite{anthropic2026claudeopus46}
          & 9.36  & 27.45 & 29.07
          & 0.06  & 6.07  & 7.96
          & 0.04  & 0.27  & 3.08   & 8.26 \\
        & GPT-5.4~\cite{openai_gpt54_2026}
          & 11.63 & \underline{35.17} & 35.55
          & 0.09  & \textbf{13.63}  & \textbf{15.46}
          & 0.02  & 0.46  & 1.92   & \underline{10.78} \\
        \hline

        \multirow{7}{*}{\shortstack[l]{Open-Source\\Spatiotemporal}}
        & EgoMask~\cite{liang2025finegrained}
          & 12.09 & 11.97 & 14.78
          & 0.03  & 3.01  & 4.17
          & 0.00  & 0.00  & 0.42   & 5.25 \\
        & Qwen2.5-VL~\cite{qwen2.5-VL}
          & 5.20  & 16.30 & 22.30
          & 0.03  & 0.40  & 2.91
          & 0.00  & 1.16  & 1.53   & 5.33 \\
        & GLM-4.6V~\cite{vteam2025glm45vglm41vthinkingversatilemultimodal}
          & 0.73  & 16.05 & 20.76
          & 0.00  & 2.57  & 5.52
          & 0.00  & 0.26  & 1.32   & 4.72 \\
        & InternVL3.5~\cite{wang2025internvl3_5}
          & 0.99  & 9.42  & 14.94
          & 0.01  & 3.21  & 5.49
          & 0.01  & 0.69  & 2.25   & 3.95 \\
        & Qwen3-VL~\cite{qwen3technicalreport}
          & 2.46  & 0.00  & 1.19
          & 0.09  & 0.00  & 0.57
          & 0.02  & 0.00  & 0.39   & 0.79 \\
        & \textbf{Ours}
          & \textbf{22.73} & 33.37          & \underline{39.27}
          & \underline{0.60}  & 6.41 & 10.48
          & \textbf{0.43}  & \underline{4.17} & \textbf{7.62}  & \textbf{13.52} \\
        \hline

        \specialrule{1.5pt}{0pt}{0pt}
    \end{tabular}}
    \caption{Unified comparison across spatial and temporal grounding subtasks on OVIS, MOT17, and MOT20
    test sets. HOTA measures spatial grounding quality; R1@0.5 and mIoU measure temporal grounding
    quality. m-HIoU is the unified joint metric computed as the mean of HOTA and mIoU averaged across
    all three datasets, and serves as the primary ranking metric. ``--'' indicates the method addresses
    only one subtask or lacks both scores required for m-HIoU. \textbf{Bold} and \underline{underline}
    indicate the best and second-best results per column. Full per-metric breakdowns are provided in
    the supplementary material.}
    \label{tab:unified_results}
    \vspace{-2mm}
\end{table*}

\subsection{Quantitative Results}
Table~\ref{tab:unified_results} reports performance across OVIS~\cite{
qi2022occluded}, MOT17~\cite{dendorfer2021motchallenge}, and
MOT20~\cite{dendorfer2020mot20}, and delivers a sobering verdict on the
state of current AI systems: no existing model family -- specialist or
general-purpose -- can reliably answer who is acting, where they are, and
when the action occurs from a single natural language query. This is
precisely the joint perceptual capability that embodied AI systems require,
and SVAG-Bench makes that gap measurable for the first time.

Specialist temporal models such as LD-DETR~\cite{zhao2025ld} and
$\rm R^2$-Tuning~\cite{liu2024r} achieve competitive mIoU on short OVIS
clips but collapse on MOT17 and MOT20, where videos span hundreds to
thousands of frames. Specialist spatial trackers such as
DKGTrack~\cite{li2025language} and TransRMOT~\cite{wu2023referring}
face the complementary problem: spatially precise tracks with no awareness
of when the queried action is actually occurring. Neither family can
produce a joint m-HIoU score at all, confirming that isolated subtask
optimization does not compose into the holistic reasoning SVAG demands.

\begin{figure}[!t]
  \begin{center}
    \includegraphics[width=0.9\linewidth]{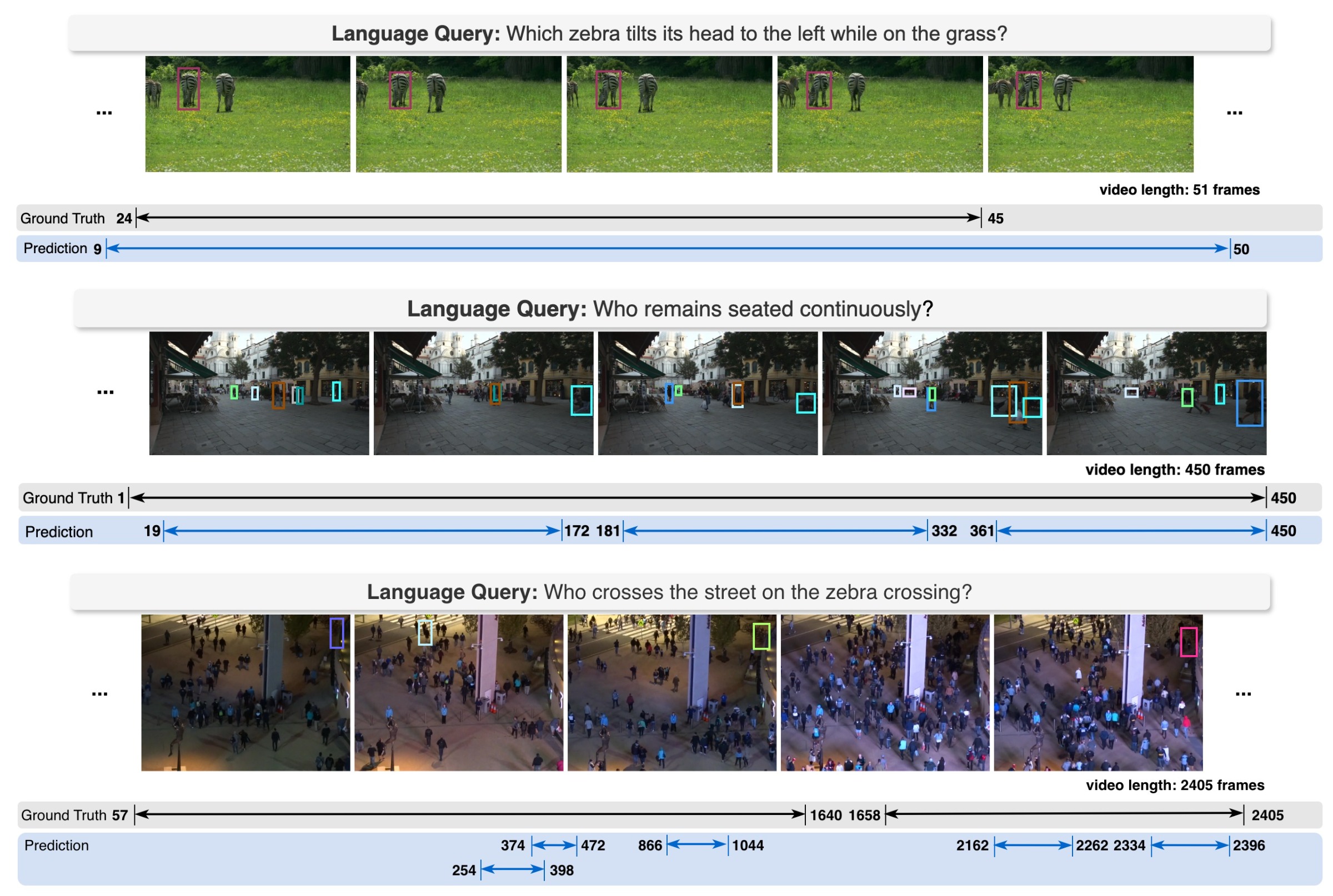}
  \end{center}
  \caption{Qualitative examples for each subdataset. From top to bottom: OVIS, MOT17, MOT20. For OVIS,  the zebra is performing a fine-grained action: tilting its head to the left across the grass. The object can be localized, even with subtle action. Detections not labeled but satisfying the query will be marked as false positives, leading to worse performance on sparse annotations in the crowd scenes. For MOT, best temporal predictions often nearly cover the full time range of the ground-truth, reflected in metrics R5 and R10, but perform poorly on R1.
    \label{fig:svag_main}
  }\vspace{-2em}
\end{figure}

LVLMs represent the current frontier of general-purpose video
understanding, yet their results expose a deeper, more consequential
failure. Both proprietary models (GPT-5.4: 10.78 m-HIoU,
Claude-Sonnet-4.6: 8.26) and the strongest open-source spatiotemporal
models (best: 5.33) achieve uniformly poor joint scores, and the cause
is not scale -- it is architecture. Current LVLMs operate as
frame-sampled visual analyzers: they compress video into sparse tokens,
discard fine-grained positional information, and produce grounding outputs
as approximate natural language rather than structured, identity-consistent
trajectories~\cite{huang2024vtimellm, wu2023referring}. This means their
spatial and temporal outputs are not jointly grounded in the same actor
identity, and moderate subtask performance on short clips does not
compose into reliable joint grounding on long, multi-actor
sequences. The pattern is consistent across every LVLM
evaluated: Qwen2.5-VL, and GLM-4.6V each achieve reasonable
R1@0.5 on OVIS but near-zero HOTA across all datasets, confirming that
their temporal estimates are not anchored to spatially coherent actor
tracks. This is not a data or scale problem that more training will
resolve; it reflects a fundamental gap in how current AI systems represent
and reason about the dynamic, multi-agent structure of real-world
video.

SVAGFormer directly addresses this gap through temporal-first gating,
constraining spatial tracking to the predicted action window and avoiding
the accumulation of false positive tracks from frames where the queried
action is absent. The result is an m-HIoU of \textbf{13.52}, nearly
double that of the best proprietary LVLM and surpassing all open-source
competitors by a substantial margin, while simultaneously achieving the
highest HOTA on OVIS (22.73) and competitive temporal scores across all
three datasets. Qualitative results are visualized in
Fig.~\ref{fig:svag_main}. Further results are in Appendix~\ref{appendix:experiments}.

\section{Conclusion, Limitations and Future Work}
\label{sec:conclusion}
We introduced SVAG on the premise that for AI to graduate from
Pattern recognition to genuine world understanding, it must perceive
actions as situated events, bound to specific agents, grounded in
space, and unfolding across time. Our results reveal that this remains
a profound open challenge, not because individual models fall short,
but because benchmarks that fragment perception into isolated sub-tasks
inevitably produce models that fragment their understanding of the world.
We offer SVAG-Bench not as a solution, but as a mirror held up to that
gap, and a step toward AI that truly comprehends the dynamic,
multi-agent world it inhabits. The benchmark's current scope is
constrained to three tracking domains, its modular pipeline precludes
joint spatio-temporal reasoning, and performance on dense long-form
videos remains near zero, limitations that point directly to the most
pressing directions ahead: broader domain coverage, end-to-end
action-grounded architectures, and validation in embodied systems.
% \newpage

\newpage
\begin{center}
    {\LARGE \textbf{Supplementary Material}}\\[0.5em]
    {\large SVAG-Bench: A Large-Scale Benchmark for \\
    Multi-Instance Spatio-temporal Video Action Grounding}
\end{center}
\vspace{1em}

\appendix
\section{Dataset Annotation Details}
\label{appendix:annotation}
\subsection{Annotation Guidelines} Human annotators follow these criteria:
\begin{enumerate}
    \item Descriptions are simple, concise, and easy to understand.
    \item Each object ID appears in at least one query.
    \item Queries define the target object without using words such as ``track'' or ``find''.
\end{enumerate}

\subsection{Language Enrichment and Example Annotations} Alternative expressions for each query are generated using GPT-3.5 to increase linguistic diversity. All annotations are verified for spelling and consistency.

Several annotation examples are shown in Table~\ref{tab:annotation}, highlighting cases where multiple objects fulfill the same query.

\begin{table}[h]
	\centering
	\renewcommand{\arraystretch}{1.1}
	\scalebox{0.72}{
		\begin{tabular}{ccccc}
			\specialrule{1.5pt}{0pt}{0pt}
			%\rowcolor{mygray} 
			Video & Language Query & Track IDs & Start Frame & End Frame \\ \hline\hline
			MOT17-02 & Woman is touching the back shoulder of another woman with her right hand & 23 & 431 & 462 \\
			 MOT17-02 & Person coming towards the camera is going through two buildings & 39 & 1 & 376 \\
             MOT17-04 & A person is keeping his right foot on a road barrier in the sidewalk & 1 & 578 & 1050 \\
             MOT17-09 & A lady is picking up the cap of the bottle with her right hand & 22 & 248 & 330 \\
             \rowcolor{red!20} MOT20-02 & Person is dancing in the open area & 140 & 2557 & 2772 \\
             \rowcolor{red!20} MOT20-02 & Person is dancing in the open area & 145 & 2543 & 2782 \\
MOT20-02 & Person in the open area is running towards the right side & 52 & 1861 & 1966 \\
\rowcolor{red!20} MOT20-02 & Person in the open area is giving high-five & 175 & 817 & 850 \\
\rowcolor{red!20} MOT20-02 & Person in the open area is giving high-five & 295 & 817 & 850 \\
MOT20-03 & A person starts running & 61 & 503 & 549 \\
MOT20-05 & A person turns forward and starts talking with another person & 793 & 1770 & 2157 \\
			  \specialrule{1.5pt}{0pt}{0pt}
	       \end{tabular}}
	\caption{Some annotation examples. Red highlights mean multiple objects under the same query.}
	\label{tab:annotation}
	\vspace{-3mm}
\end{table}

\section{Dataset Statistics Details}
\label{appendix:statistics}
\subsection{Per-subset Statistics} Table~\ref{tab:dataset_subsets} shows detailed statistics for each source subset.

\begin{table}[h]
	\centering
	\renewcommand{\arraystretch}{1.1}
	\scalebox{0.82}{
		\begin{tabular}{lccccc}
			\specialrule{1.5pt}{0pt}{0pt}
			%\rowcolor{mygray} 
			Dataset &       & Videos & Queries & Tracks & Distinct Verbs (Action) \\\hline\hline
\multirow{3}{*}{OVIS~\cite{qi2022occluded}}
& Train & 533 & 7352 & 3685 & 703 \\
& Test\textsuperscript{\dag}  & 137 & 1356 & 678 & 273 \\
\cmidrule{2-6}
& \textbf{Total} & 670 & 8708 & 4363 & 768 \\\hline
\multirow{3}{*}{MOT17~\cite{dendorfer2021motchallenge}}
& Train & 7 & 1538 & 775 & 216 \\
& Test & 7 & 1397 & 698 & 129 \\
\cmidrule{2-6}
& \textbf{Total} & 14 & 2935 & 1473 & 263 \\
\hline
\multirow{3}{*}{MOT20~\cite{dendorfer2020mot20}} 
& Train & 2 & 4544 & 2275 & 103 \\
& Test\textsuperscript{*}  & 2 & 3403 & 1670 & 135 \\
\cmidrule{2-6}
& \textbf{Total} & 4 & 7947 & 3945 & 214 \\\midrule 
\textbf{SVAG-Bench} & \textbf{Total} & 688 & 19590 & 9781 & 903 \\
  \specialrule{1.5pt}{0pt}{0pt}
	       \end{tabular}}
	\caption{Statistics of SVAG-Bench's settings. Dataset marked with \dag\ originates from the OVIS valid set. Dataset marked with * originates from the MOT20 train set, but we split it into training and test sets for our experiments.}
	\label{tab:dataset_subsets}
	\vspace{-3mm}
\end{table}

\subsection{Distinct Verbs (Action)} 
\begin{figure}
  \begin{center}
    \includegraphics[width=1\linewidth]{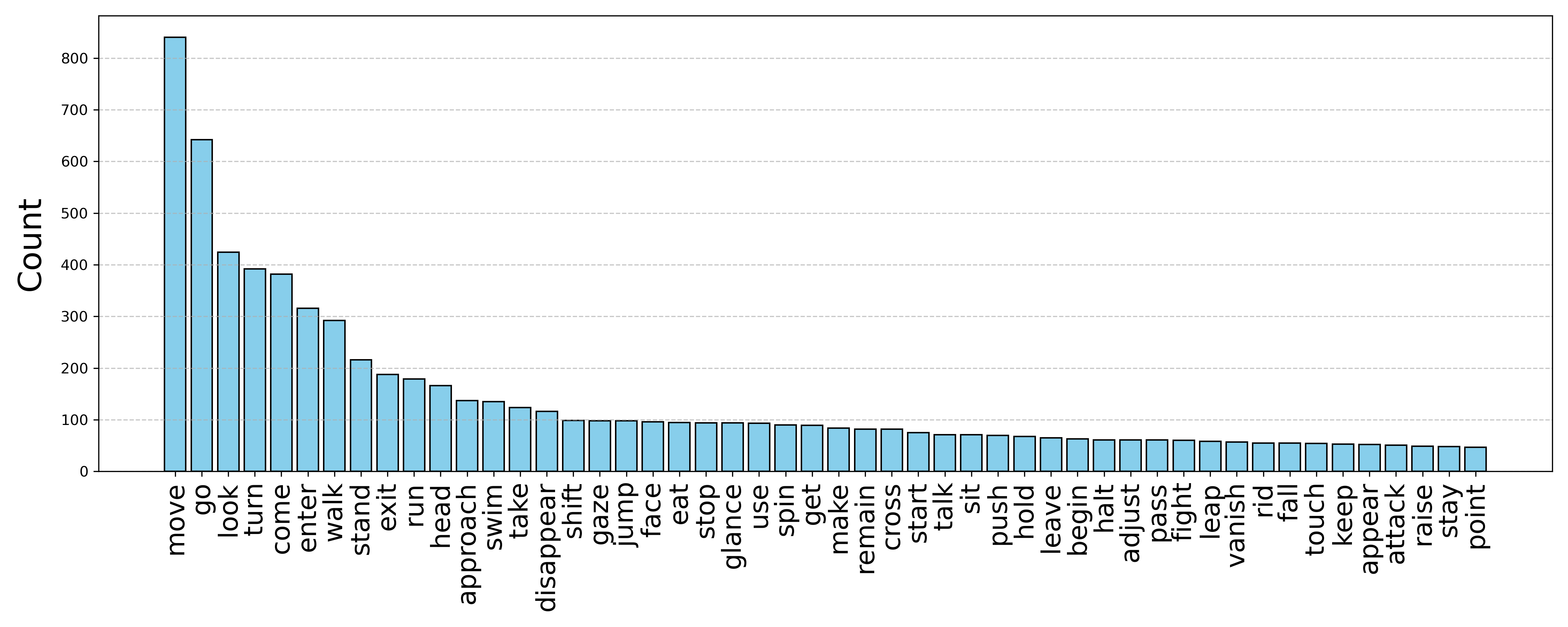}
  \end{center}
  \caption{Distribution of the most common 50 actions. The action distribution is strongly skewed, with a few general motion verbs (e.g., move, go, look) appearing far more frequently than others. Most actions occur less than 300 times, revealing a long-tail pattern of  SVAG-Bench datasets.
    \label{fig:top50}
  }
\end{figure}

The distribution of the top 50 most frequent base-form verbs (actions) is illustrated in Fig.~\ref{fig:top50}. The most commonly annotated actions in SVAG-Bench include ``move'' (840 instances), ``go'' (642), ``look'' (424), ``turn'' (392), and ``come'' (382). This distribution reflects different dynamic behaviors across various object categories and domains.

\subsection{Video Length} Table~\ref{tab:video_length} reports the average video duration of the three source datasets (MOT17, MOT20, OVIS). Videos from MOT20 are the longest on average, while OVIS provides the shortest but more diverse videos with rich object interactions.
\begin{table}[h]
	\centering
	\renewcommand{\arraystretch}{1.1}
	\scalebox{0.72}{
		\begin{tabular}{lcccc}
			\specialrule{1.5pt}{0pt}{0pt}
			%\rowcolor{mygray} 
			Dataset & Videos & avg. Video Length & avg. Video Length (training set) & avg. Video Length (test set) \\ \hline\hline
			OVIS & 670 & 67.7 & 69.4 & 61.1 \\
MOT17 & 14 & 802.5 & 759.4 & 845.6 \\
MOT20 & 4 & 2232.8 & 1605.5 & 2860.0 \\ \specialrule{1.5pt}{0pt}{0pt}
	       \end{tabular}}
	\caption{Comparison of video lengths in different datasets. The video length of MOT20 is the longest, while OVIS is the shortest.}
	\label{tab:video_length}
	\vspace{-3mm}
\end{table}

\subsection{Object Category Distribution}
\begin{figure}[htbp]
  \begin{center}
    \includegraphics[width=0.7\linewidth]{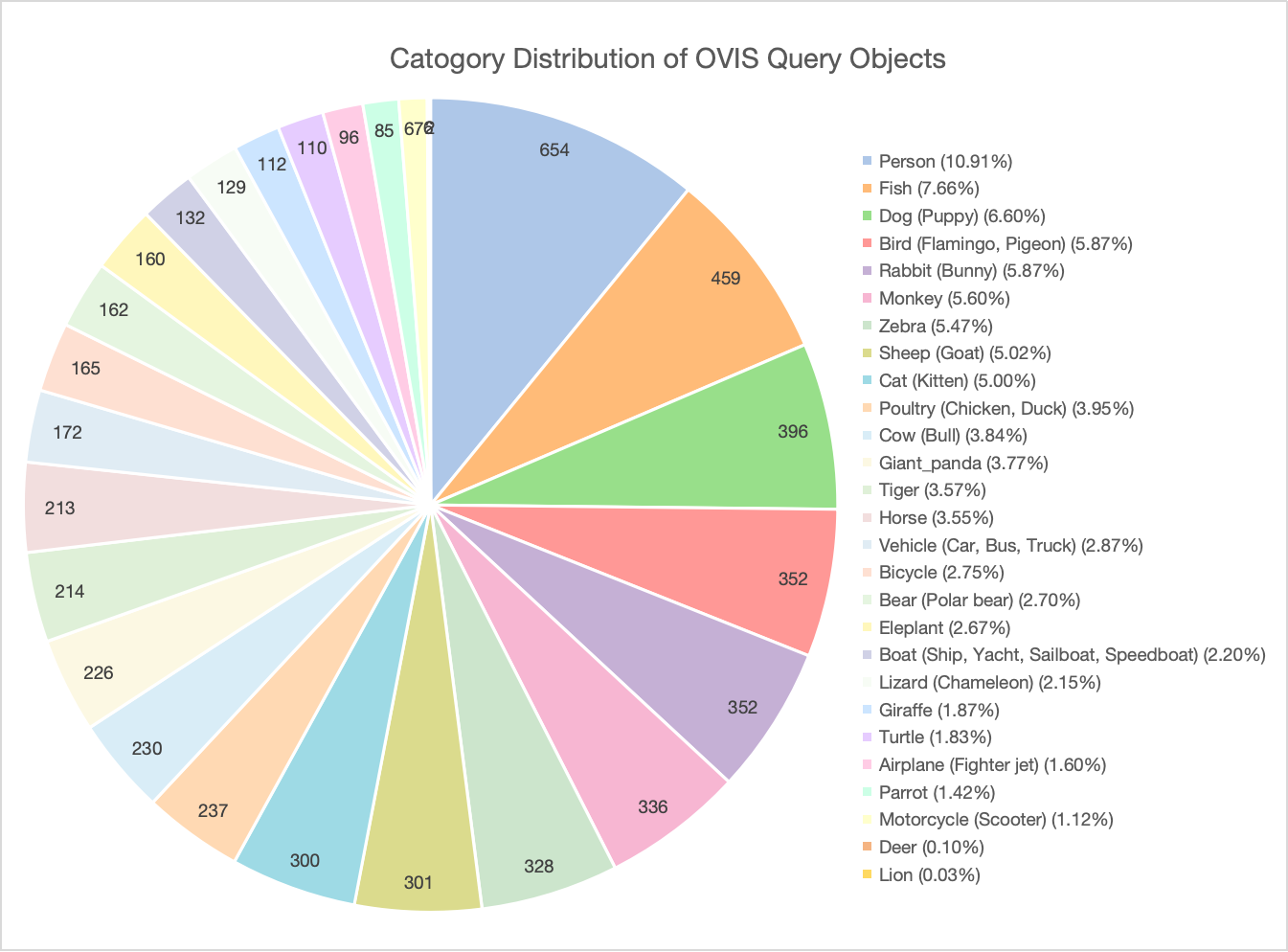}
  \end{center}
  \caption{Category distribution of OVIS query objects. OVIS includes diverse referent object categories such as Person, Fish, Dog, Bird, and Rabbit. This broad coverage enables SVAG-Bench to support multi-object, human-centric, and animal behavior grounding tasks.}
    \label{fig:category_ovis}
\end{figure}

Another distinguishing feature of SVAG-Bench is its breadth of referent object categories. As shown in Fig.~\ref{fig:category_ovis}, OVIS provides a wide variety of categories (e.g., fish, dogs, birds, rabbits, monkeys), while MOT17 and MOT20 are dominated by the ``Person'' class (over 90\%). 
This diversity allows SVAG-Bench to support human-centric, multi-object, and animal behavior grounding tasks.

\section{Data Pre-processing}
\label{appendix:preprocess} 
This section explains how we preprocessed the original dataset for subsequent spatial and temporal localization models.

\textbf{Spatial Grounding.} We cannot obtain the official bounding box information of the OVIS validation set, which is needed for the downstream spatial grounding task. Therefore, we first apply video instance segmentation with DVIS++~\cite{zhang2023dvisplus} on the OVIS validation set to obtain object masks and convert them into bounding boxes. DVIS++ obtains the leading performance on OVIS. Table~\ref{tab:dvis++} compares our re-implementation with the original, achieving similar AP (52.44 vs 53.4). Since DVIS++ outputs multiple instances with occasional ID switches, we apply lightweight manual post-processing to ensure the necessary number of instances and stable object trajectories. These bounding boxes and language queries are then reformatted to meet the input requirements of downstream models.  

\begin{table}[h]
	\centering
	\renewcommand{\arraystretch}{1.1}
	\scalebox{0.92}{
		\begin{tabular}{lcccccc}
			\specialrule{1.5pt}{0pt}{0pt}
			%\rowcolor{mygray} 
			Method & Backbone & AP & $\mathrm{AP}_{50}$ & $\mathrm{AP}_{75}$ & $\mathrm{AR}_{1}$ & $\mathrm{AR}_{10}$ \\ \hline\hline
			DVIS++~\cite{zhang2023dvisplus} & VIT-L & 53.4 & 78.9 & 58.5 & 21.1 & 58.7 \\\hline
\textbf{DVIS++}~\cite{zhang2023dvisplus} (Ours) & VIT-L & 52.44 & 77.11 & 57.41 & 21.03 & 57.85 \\ \specialrule{1.5pt}{0pt}{0pt}
	       \end{tabular}}
	\caption{Comparison of DVIS++ results between ours (AP 52.44) and the official (AP 53.4) on OVIS}
	\label{tab:dvis++}
	\vspace{-3mm}
\end{table}

\textbf{Temporal Grounding.} As each query in QVHighlights only has one object that meets the description, FlashVTG cannot perform temporal grounding for multiple instances under the same query. Therefore, we merge the time intervals for different instances executing the same query to ensure that each query is input only once.

To evaluate individual model performances, we use bounding-box ground truth for the spatial grounding model and temporal-interval ground truth for the temporal grounding model.

% \section{SVAGFormer Pipeline \textcolor{red}{[Keep or Remove this section?]}}
% The pipeline is visualized in Fig. \ref{fig:SVAGFormer}.

% \begin{figure}[htbp]
%   \begin{center}
%     \includegraphics[width=0.76\linewidth]{figures/pipeline1.jpg}
%   \end{center}
%   \caption{\tanveer{please follow accepted papers on how they create method figure, this is not suitable for publication}Illustration of the components of SVAGFormer Pipeline}
%       \label{fig:SVAGFormer}
% \end{figure}

\section{Experiments Details}
\label{appendix:experiments}

\subsection{Evaluation Metrics} 
\label{appendix:metrics}
\textbf{Spatial Grounding Evaluation.} For spatial grounding, we utilize the Higher Order Tracking Accuracy (HOTA) metric~\cite{luiten2021hota}, which has gained recognition as a robust and unified evaluation standard in the multi-object tracking (MOT) community. HOTA is designed to jointly measure both detection accuracy and association quality within a single coherent framework. Specifically, it computes the geometric mean of two sub-metrics: Detection Accuracy (DetA) and Association Accuracy (AssA), as follows~\cite{wu2023referring}:
\begin{equation}
\text{HOTA} = \sqrt{\text{DetA} \cdot \text{AssA}}
\end{equation}

DetA reflects how well the tracker detects objects of interest, while AssA measures how accurately the tracker maintains the identities of these objects over time. By combining these two dimensions, HOTA ensures a balanced evaluation of the tracking quality~\cite{luiten2021hota}.

In the context of referring multi-object tracking (RMOT), as introduced in~\cite{wu2023referring}, HOTA is modified to suit the specific requirements of the task. In particular, predicted tracks that correspond to visible objects not referenced by any sentence query are treated as false positives. This modification ensures that the evaluation focuses strictly on the ability of the model to identify and track only objects relevant to the natural language query, thereby better aligning with the goal of language-guided spatial grounding. The overall HOTA is obtained by first computing per query and then averaging across all sentence queries in the dataset~\cite{wu2023referring}.

\textbf{Temporal Grounding Evaluation.} For the evaluation of temporal grounding, we follow prior works~\cite{Cao_2025_WACV, jiang2024prior, lei2021detecting} and adopt a set of standard metrics for the moment retrieval task, namely Recall at 1 (R1@X), mean Average Precision (mAP), and mean Intersection over Union (mIoU).

R1@X measures the ability of the model to correctly localize the most relevant video moment for a given query. Specifically, the model chooses the temporal segment prediction with the highest confidence score, and the prediction is considered correct if its Intersection-over-Union (IoU) with the ground truth segment is greater than or equal to a predefined threshold X~\cite{Cao_2025_WACV}. This metric is crucial for evaluating retrieval accuracy in a practical scenario where only the top prediction is used.

To more thoroughly assess the retrieval capability of the model, we also report R5@X and R10@X, which extend the recall evaluation to the top-5 and top-10 predicted moments, respectively. These metrics evaluate whether any of the top-k predicted segments have sufficient overlap with the ground truth. They are useful for analyzing the model's performance in applications where multiple candidate predictions are considered.

mAP is computed over multiple IoU thresholds and measures the model's ability to retrieve accurate temporal segments given a query. mIoU measures the average overlap between the predicted and ground truth moments across all queries, reflecting the temporal localization accuracy of the model~\cite{Cao_2025_WACV}.

For all recall-based metrics, we report results at three commonly used IoU thresholds: X $\in$ \{0.1, 0.3, 0.5\}, which represent different levels of localization strictness. This multi-threshold reporting allows us to analyze the model's performance under both lenient and stringent evaluation conditions.

\subsection{Extended Implementation Details}
\label{appendix:implementation}

\textbf{Spatial Grounding.} Table~\ref{tab:spatial_runtime} reports the training runtime for spatial grounding on different datasets.
\begin{table}[h]
	\centering
	\renewcommand{\arraystretch}{1.1}
	\scalebox{0.82}{
		\begin{tabular}{lccc}
			\specialrule{1.5pt}{0pt}{0pt}
			%\rowcolor{mygray} 
			Dataset & Epochs & Runtime & GPUs \\ \hline\hline
OVIS & 60 & ~8d13h & 4 \\
MOT17 & 60 & ~5d6h & 4 \\
OVIS (w/o rk2)  & 5 & ~1d10.5h & 4 \\
MOT17 (w/o rk2)  & 5 & ~5.5h & 4 \\
MOT20 (w/o rk2)  & 5 & ~2h50m & 4 \\
 \specialrule{1.5pt}{0pt}{0pt}
	       \end{tabular}}
	\caption{Training runtime for spatial grounding on different datasets}
	\label{tab:spatial_runtime}
	\vspace{-3mm}
\end{table}

\textbf{Temporal Grounding Settings.} Since the model does not perform multi-instance localization, we preprocessed the dataset. For different instances of the same query, we first merged their occurrence time intervals and then generated annotations. Table~\ref{tab:temporal_para} summarizes the hyperparameters, and Table~\ref{tab:temporal_runtime} shows the training runtime for temporal grounding across datasets.

\begin{table}[ht]
	\centering
	\renewcommand{\arraystretch}{1.1}
	\scalebox{0.72}{
		\begin{tabular}{lc}
			\specialrule{1.5pt}{0pt}{0pt}
			%\rowcolor{mygray} 
			Parameter & Value \\ \hline\hline
max\_v\_l & 250 (OVIS), 750 (MOT17), 1658 (MOT20) \\
Buffer size & 1024 (OVIS/MOT17), 2048 (MOT20) \\
 \specialrule{1.5pt}{0pt}{0pt}
	       \end{tabular}}
	\caption{Hyperparameters for temporal grounding}
	\label{tab:temporal_para}
	\vspace{-3mm}
\end{table}

\begin{table}[ht]
	\centering
	\renewcommand{\arraystretch}{1.1}
	\scalebox{0.72}{
		\begin{tabular}{lccc}
			\specialrule{1.5pt}{0pt}{0pt}
			%\rowcolor{mygray} 
			Dataset & Epochs & Runtime & GPUs \\ \hline\hline
OVIS & 150 & 1h19m & 1xA40 \\
MOT17 & 150 & ~0.5h & 1xA40 \\
MOT20  & 150 & ~1h5m & 1xA100-SXM4-80GB \\
 \specialrule{1.5pt}{0pt}{0pt}
	       \end{tabular}}
	\caption{Training runtime for temporal grounding on different datasets}
	\label{tab:temporal_runtime}
	\vspace{-3mm}
\end{table}

\subsection{Ablation Study}
\label{appendix:ablation}

\textbf{(1) Pretrained weight from Refer-KITTI-V2.} 
% \paragraph{Setup.}
We assessed the effect of initializing TempRMOT with pretrained weights from Refer-KITTI-V2 (checkpoint\_rk2). For this experiment, we fine-tuned the model for 5 epochs: the learning rate was kept constant for the first 3 epochs and decayed by a factor of 10 from epoch 4 onward; all other training hyperparameters remained unchanged. Results are reported in Table~\ref{tab:rk2}.

% Key observations compared to Table~\ref{tab:performance_tmeprmot} are:
\begin{itemize}
  \item Pretraining substantially improves association accuracy across datasets. In particular, AssA increases from 46.625 to 51.617 on OVIS, and from 9.171 to 20.068 on MOT17.
  \item HOTA increases by approximately 2\% on OVIS and MOT17. For MOT20, HOTA slightly decreases (0.428 $\rightarrow$ 0.333).
  \item Although HOTA drops slightly on MOT20,  AssA rises markedly (2.971 $\rightarrow$ 11.525); this suggests that pretraining improves temporal association but does not fully adapt the detection component to MOT20's crowding and small-object scenario.
\end{itemize}

Overall, checkpoint\_rk2 provides task-aligned representations that strengthen identity association across frames. The gains in AssA indicate more robust temporal linking of identities, which is particularly beneficial in occluded or interaction-rich scenarios (e.g., OVIS). The MOT20 behaviour highlights a detection-association trade-off: pretraining enhances association capabilities but may require detection finetuning.

\begin{table}[h]
	\centering
	\renewcommand{\arraystretch}{1.1}
	\scalebox{0.72}{
		\begin{tabular}{lcccccccc}
			\specialrule{1.5pt}{0pt}{0pt}
			%\rowcolor{mygray} 
			Dataset & HOTA & DetA & AssA & DetRe & DetPr & AssRe & AssPr & LocA \\ \hline\hline
			OVIS & 24.611 & 11.886 & 51.617 & 20.193 & 21.829 & 64.761 & 65.701 & 82.517 \\
MOT17 & 2.4957 & 0.34994 & 20.068 & 0.56698 & 0.90229 & 27.155 & 37.495 & 70.787 \\
MOT20 & 0.33277 & 0.011678 & 11.525 & 0.51524 & 0.011947 & 22.567 & 21.443 & 65.704  \\ \specialrule{1.5pt}{0pt}{0pt}
	       \end{tabular}}
	\caption{Performance on the different datasets using TempRMOT~\cite{zhang2024bootstrapping} with weight from rk2. HOTA increases by approximately 2\% on OVIS and MOT17. Pretraining substantially improves association accuracy across dataset.}
	\label{tab:rk2}
	\vspace{-3mm}
\end{table}

% To investigate the impact of certain parameters on spatial grounding model performance, we conducted ablation studies on the OVIS dataset, focusing specifically on two key parameters: \textbf{history length} and \textbf{referring threshold}.

Following this experiment, we investigated the effect of \textbf{history length} and \textbf{referring threshold} on OVIS on spatial grounding model performance.

\textbf{(2) Number of Temporal Memory.} 
We studied the effect of temporal memory length on model performance during inference. Training memory length was fixed at 5, while inference memory length was varied among $\{5, 6, 8\}$ frames. Results are summarized in Table~\ref{tab:frame_length}.

% Key observations are:
\begin{itemize}
  \item Increasing memory length from 5 to 8 improves HOTA, DetA, and AssA, indicating that a longer temporal context helps capture dynamic changes and object interactions more effectively.
  \item Performance at memory length 6 is slightly worse than at 5, particularly for AssA, showing that the gains are not strictly monotonic.
\end{itemize}

While a longer memory window typically provides richer temporal information, the improvements may saturate or fluctuate depending on dataset characteristics. Choosing memory length requires balancing richer temporal context with computational overhead. Short windows miss important historical context, while overly long ones may add redundancy. Therefore, memory length should be tuned to dataset complexity and runtime constraints.

% To investigate the effect of memory length on performance during the inference phase, the memory length during training was fixed at 5 and varied during inference. Specifically, we tested inference memory lengths of 5, 6, and 8 frames.

% As shown in Table \ref{tab:frame_length}, increasing the memory length from 5 to 8 leads to improvements across all evaluation metrics, including HOTA, DetA, and AssA. This suggests that leveraging longer temporal context allows the model to better capture the dynamic changes and object interactions in the video, thus enhancing overall performance.

% However, the performance improvement is not strictly monotonic. Notably, the model's performance at length 6 is slightly worse than at length 5, especially on AssA. This observation implies that although aggregating more historical information generally benefits the model, the benefit may saturate or fluctuate depending on the specific memory length and dataset characteristics.

\begin{table}[h]
	\centering
	\renewcommand{\arraystretch}{1.1}
	\scalebox{0.82}{
		\begin{tabular}{cccc}
			\specialrule{1.5pt}{0pt}{0pt}
			% \rowcolor{mygray} 
			Length & HOTA & DetA & AssA \\ \hline\hline
			5 & 24.214 & 11.608 & 51.086 \\
6 & 23.769 & 11.447 & 49.98 \\
\textbf{8} & \textbf{24.611} & \textbf{11.886} & \textbf{51.617} \\ \specialrule{1.5pt}{0pt}{0pt}
	       \end{tabular}}
	\caption{Different lengths for inference on OVIS. Increasing memory length from 5 to 8 improves HOTA, DetA, and AssA. However, the gains are not strictly monotonic.}
	\label{tab:frame_length}
	\vspace{-3mm}
\end{table}

% Overall, these findings highlight the importance of carefully selecting the temporal memory length in practical applications. A memory length that is too short may fail to capture sufficient historical context. However, an excessively long memory may introduce redundancy and increase computational overhead, necessitating a balance between the two.
%To investigate the impact of memory length during the inference phase, we fixed the training memory length to 5 and varied it in the inference phase. As shown in Table \ref{tab:frame_length}, while increasing the memory length to 8 leads to improved performance in all metrics, the trend is not strictly monotonic - length 6 performs slightly worse than length 5. This suggests that although aggregating more historical information can generally enhance model performance, the benefit may saturate or fluctuate depending on the specific memory length and dataset characteristics.

\textbf{(3) Referring Threshold.} 
The referring threshold is used to filter candidate objects against the language query. We varied the threshold from 0.1 to 0.5 (step 0.1) on OVIS and MOT17 (with rk2 pretraining). The score threshold was fixed at 0.4 for all MOT17 experiments. Results are summarized in Table~\ref{tab:refer_threshold} and Table~\ref{tab:refer_threshold_mot17}.

\begin{itemize}
  \item On OVIS, the best HOTA (24.61) is achieved at threshold 0.3, with moderate fluctuations between 0.2 and 0.4.  
  \item On MOT17, lowering the threshold to 0.1 yields the highest HOTA (4.04).  
\end{itemize}

The results highlight a precision–recall trade-off: Higher thresholds improve precision but reduce recall, discarding potentially useful detections. Lower thresholds retain more candidates, boosting recall and overall HOTA in challenging datasets like MOT17. Optimal referring thresholds are dataset-dependent: OVIS benefits from a stricter threshold (0.3), while MOT17 requires a more permissive setting (0.1). This demonstrates the importance of threshold tuning for balancing recall and precision across different scenarios.

\begin{table}[h]
	\centering
	\renewcommand{\arraystretch}{1.1}
	\scalebox{0.72}{
		\begin{tabular}{cccccc}
			\specialrule{1.5pt}{0pt}{0pt}
			% \rowcolor{mygray} 
			Referring threshold & HOTA & DetA & AssA & DetRe & DetPr \\ \hline\hline
			0.1 & 20.111 & 8.7548 & 46.865 & \textbf{46.762} & 9.6028 \\
0.2 & 21.452 & 9.7768 & 47.724 & 37.108 & 11.554 \\
0.3 & \textbf{24.611} & \textbf{11.886} & 51.617 & 20.193 & 21.829 \\
0.4 & 21.357 & 8.4915 & \textbf{54.325} & 9.8851 & 36.113 \\
0.5 & 15.3 & 4.4273 & 53.518 & 4.625 & \textbf{48.7}\\ \specialrule{1.5pt}{0pt}{0pt}
	       \end{tabular}}
	\caption{Different referring threshold $\beta_{\mathrm{ref}}$ on OVIS. The best HOTA (24.61) is achieved at threshold 0.3, with moderate fluctuations
between 0.2 and 0.4.}
	\label{tab:refer_threshold}
	\vspace{-3mm}
\end{table}

%Due to the poor results on the MOT dataset, we also applied different thresholds to the MOT17 checkpoint trained using rk2 as pre-trained weights. Since the current threshold is 0.3, we chose the same threshold set as OVIS, which is 0.1 to 0.5 with an increment of 0.1 while keeping the score threshold fixed at 0.4. The results are shown in Table \ref{tab:refer_threshold_mot17}, where we can see that the HOTA is reduced by 0.48 at 0.2, but improved by 1.54 at 0.1 compared to 2.4957. We observe a clear precision-recall trade-off as the referring threshold varies. When the threshold is set to 0.1, the detection recall (DetRe) reaches its highest value of 4.76, indicating that more ground truth targets are successfully retrieved. However, this comes at the cost of a slightly reduced precision (DetPr = 0.87). Conversely, increasing the threshold to 0.3 leads to a lower recall (0.57), but the detection precision improves to 0.90, suggesting that predictions become more selective and accurate. At 0.4 and 0.5, DetRe and DetPr have the same trend. This trade-off is typical in detection tasks when adjusting matching thresholds.
% We extended our investigation to the MOT17 dataset, which exhibited poor performance under the default settings. For consistency, we evaluated the same threshold range (0.1 to 0.5, step 0.1) using a checkpoint trained on the rk2 pre-trained weight, keeping the score threshold fixed at 0.4. Specifically, the score threshold filters out the detections with low confidence scores. 

\begin{table}[h]
	\centering
	\renewcommand{\arraystretch}{1.1}
	\scalebox{0.72}{
		\begin{tabular}{cccccc}
			\specialrule{1.5pt}{0pt}{0pt}
			% \rowcolor{mygray} 
			Referring threshold & HOTA & DetA & AssA & DetRe & DetPr \\ \hline\hline
			0.1 & \textbf{4.0378} & \textbf{0.74358} & \textbf{23.2} & \textbf{4.7601} & 0.87094 \\
0.2 & 2.0189 & 0.34872 & 12.416 & 0.86141 & 0.58108 \\
0.3 & 2.4957 & 0.34994 & 20.068 & 0.56698 & 0.90229 \\
0.4 & 1.655 & 0.28143 & 10.455 & 0.34118 & \textbf{1.5701} \\
0.5 & 1.4319 & 0.22516 & 9.3577 & 0.26326 & 1.5225 \\ \specialrule{1.5pt}{0pt}{0pt}
	       \end{tabular}}
	\caption{Different referring threshold $\beta_{\mathrm{ref}}$ on MOT17. Lowering the threshold to 0.1 yields the highest HOTA (4.04). The performance on the threshold of 0.2 is not monotonically increasing.}
	\label{tab:refer_threshold_mot17}
	\vspace{-3mm}
\end{table}

The following two ablation studies were conducted on the temporal grounding model. 

\textbf{(4) NMS Threshold.} We compared Non-Maximum Suppression (NMS) thresholds of 0.5 and 0.7 in the FlashVTG framework, keeping all other parameters fixed. Results are shown in Table~\ref{tab:nms_flashvtg}.
\begin{itemize}
  \item NMS = 0.5 improves R5@X, R10@X, and mAP, with the largest gains on MOT17 (e.g., R10@0.5 +10.42\%).  
  \item OVIS benefits more from NMS = 0.7 in R1@X and mIoU (+2.1\% and +0.76\%, respectively).  
  \item MOT20 shows minimal differences, though NMS = 0.5 performs better on R10 and mAP.  
\end{itemize}

Lower NMS thresholds apply stronger suppression, reducing redundant predictions, while a higher value retains more overlapping results.

\begin{table*}[htb]
	\centering
	\renewcommand{\arraystretch}{1.1}
	\scalebox{0.72}{
		\begin{tabular}{lccccccccccccc}
			\specialrule{1.5pt}{0pt}{0pt}
            	& \multicolumn{3}{c}{R1} & \multicolumn{3}{c}{R5} & \multicolumn{3}{c}{R10} & \multicolumn{3}{c}{mAP} &  \\ 
\multirow{-2}{*}{Dataset}
			%\rowcolor{mygray} 
			%\multirow{-2}{*}{\cellcolor{mygray} Dataset}
            & @0.1 & @0.3 & @0.5 & @0.1 & @0.3 & @0.5 & @0.1 & @0.3 & @0.5 & @0.1 & @0.3 & @0.5 & \multirow{-2}{*}{ mIoU} \\
            % \multirow{-2}{*}{\cellcolor{mygray} mIoU} \\
			\hline
			\hline
			OVIS\textsuperscript{\S} & 80.27 & 52.55 & 33.04 & \textbf{96.9} & \textbf{80.82} & \textbf{66.19} & \textbf{99.33} & \textbf{91.91} & \textbf{80.38} & \textbf{85.98} & \textbf{63.75} & \textbf{45.84} & 38.51 \\
OVIS\textsuperscript{\dag} & \textbf{82.37} & \textbf{52.66} & \textbf{33.37} & 91.57 & 75.06 & 59.31 & 94.90 & 85.03 & 75.06 & 85.63 & 61.22 & 43.98 & \textbf{39.27} \\\midrule 
MOT17\textsuperscript{\S} & \textbf{35.67} & 12.42 & \textbf{6.81} & \textbf{80.56} & \textbf{46.49} & \textbf{23.65} & \textbf{90.58} & \textbf{57.31} & \textbf{31.66} & \textbf{51.26} & \textbf{25.06} & \textbf{11.97} & \textbf{11.51} \\
MOT17\textsuperscript{\dag} & 30.86 & \textbf{13.23} & 6.41 & 74.15 & 39.48 & 15.63 & 83.57 & 48.90 & 21.24 & 43.67 & 21.79 & 8.97 & 10.48 \\\midrule 
MOT20\textsuperscript{\S} & 20.14 & 9.95 & 4.17 & 48.15 & 22.45 & 12.73 & \textbf{65.51} & \textbf{38.43} & \textbf{22.22} & \textbf{21.53} & \textbf{10.58} & \textbf{5.66} & 7.62 \\
MOT20\textsuperscript{\dag} & 20.14 & 9.95 & 4.17 & 48.15 & 22.45 & 12.73 & 61.81 & 31.02 & 20.83 & 20.56 & 10.07 & 5.47 & 7.62 \\\specialrule{1.5pt}{0pt}{0pt}
	\end{tabular}}
	\caption{Performance on the different datasets using FlashVTG \cite{Cao_2025_WACV}. Datasets marked with \dag\ use NMS 0.7. Datasets marked with \S\ use NMS 0.5. The higher score is highlighted in bold. Lower NMS thresholds apply stronger suppression, reducing redundant predictions, leading to better performance.}
	\label{tab:nms_flashvtg}
	\vspace{-2mm}
\end{table*}

\subsection{Additional Qualitative Results}
\label{appendix:qualitative}
\begin{figure}[htbp]
  \begin{center}
    \includegraphics[width=1\linewidth]{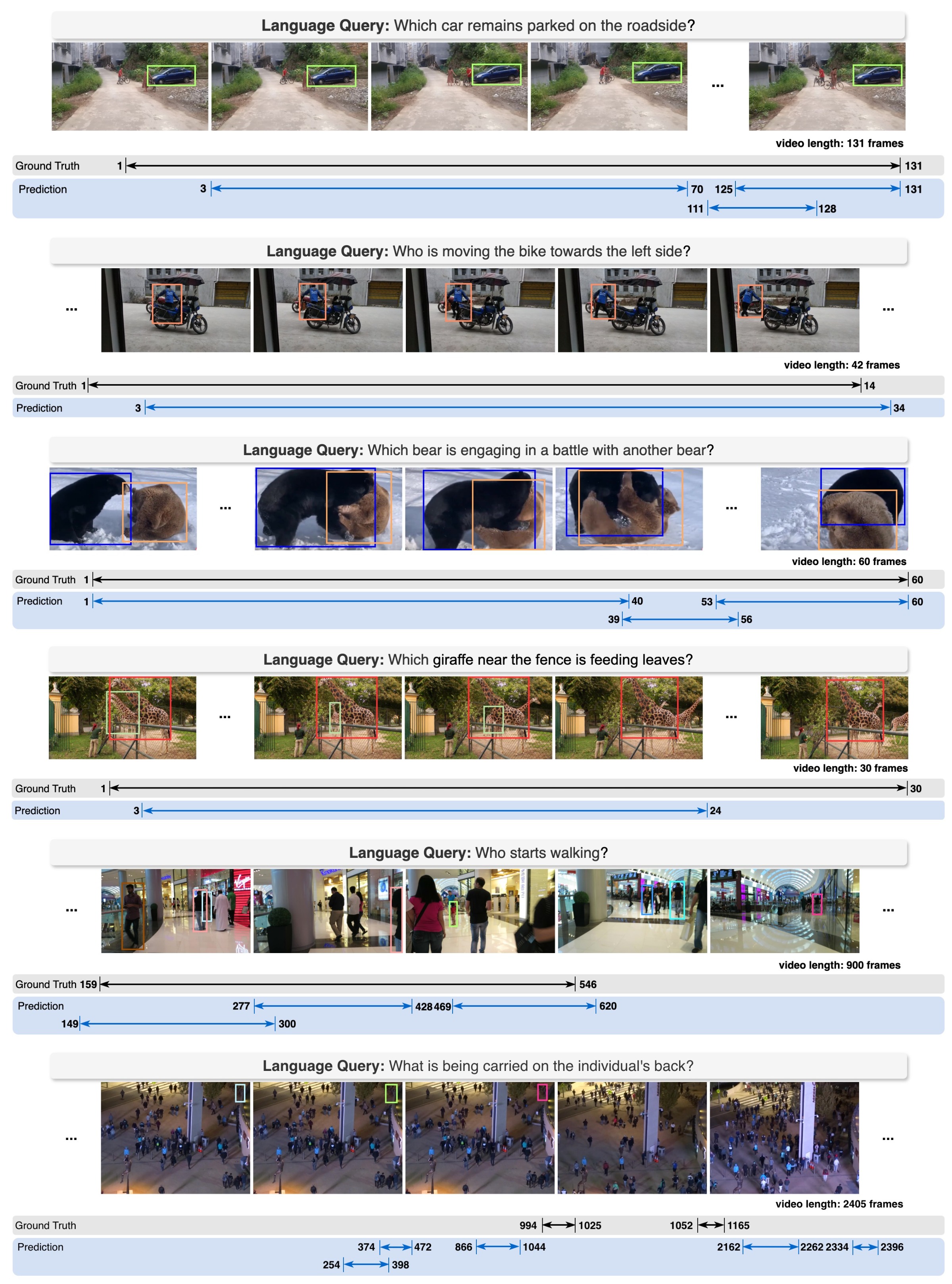}
  \end{center}
  \caption{Qualitative examples on SVAG-Bench. From top to bottom, there are four OVIS, MOT17, and MOT20. The first two examples have a single referent, and the others have multiple referents. The images show the referent's box. The blue bar shows the predicted time window(s). They cover part or all of the ground truth.
    \label{fig:svag_visual}
  }
\end{figure}

Fig.~\ref{fig:svag_visual} shows additional qualitative examples from SVAG-Bench, including OVIS, MOT17, and MOT20 scenes. These examples illustrate the model's ability to localize objects both spatially and temporally across diverse scenarios.

The predicted temporal segments typically align well with the ground-truth intervals, either fully or partially overlapping, demonstrating robust temporal grounding performance. MOT17 examples, despite dense scenes, show partial success in aligning language queries with specific individuals. However, the spatiotemporal grounding performance is poor for MOT20.
% \begin{figure}[htbp]
%   \begin{center}
%     \includegraphics[width=1\linewidth]{figures/mot17_vis.jpg}
%   \end{center}
%   \caption{Qualitative examples on MOT17. Detections not labeled but satisfying the query will be marked as false positives, leading to worse performance on sparse annotations in the crowd scenes. Best temporal predictions often nearly cover the full time range of the ground-truth, reflected in metrics R5 and R10, but perform poorly on R1.
%     \label{fig:svag_visual_mot17}
%   }
% \end{figure}
% \begin{figure}[htbp]
%   \begin{center}
%     \includegraphics[width=1\linewidth]{figures/ovis2.jpg}
%   \end{center}
%     \caption{A qualitative visualization example of a zebra performing a fine-grained action: tilting its head to the left across the grass. TempRMOT can localize the object, even with subtle action.  \label{fig:zebra_vis}}
% \end{figure}

\subsection{Extended Results Analysis}
\label{appendix:results_analysis}

Animal-based entities dominate the referents: \textit{Dog} (12 occurrences), \textit{Rabbit (Bunny)} (9), and \textit{Cow (Bull)} (7). These categories often exhibit distinctive motions (running, eating, fighting) that facilitate action-based localization. In addition, inanimate and vehicle categories (e.g., Airplane, Bicycle, Ship) also appear, typically associated with salient dynamics such as takeoff, driving, and landing.

Comparison with the full dataset distribution (Fig.~\ref{fig:category_ovis}) reveals that some rare categories (e.g., Airplane, Bear) are frequent in high-HOTA sequences, while some common ones (e.g., Fish, Poultry) are underrepresented. This indicates that semantic salience and motion diversity, rather than frequency, are key factors in successful grounding.

Fig.~\ref{fig:svag_main} shows the detection results on MOT. In crowd scenes, sparse annotations can cause evaluation bias. Specifically, detections that are not labeled but satisfy the query will be marked as false positives. Multiple temporal predictions often nearly cover the full time range of the ground-truth, suggesting temporal consistency is preserved. It is reflected in metrics R5 and R10, but performs poorly on R1.

\subsection{More Quantitative Results}
 We evaluate the performance on three datasets (OVIS \cite{qi2022occluded}, MOT17 \cite{dendorfer2021motchallenge}, and MOT20 \cite{dendorfer2020mot20}) separately using two different benchmark frameworks named TempRMOT \cite{zhang2024bootstrapping} and FlashVTG \cite{Cao_2025_WACV} based on the experimental setup mentioned above. The results are reported in Tables~\ref{tab:compar_temp_ovis}--\ref{tab:performance_flashvtg}. Models perform consistently better on OVIS than on MOT17 and MOT20 across all metrics, suggesting that these models have better generalization capabilities under complex occlusion conditions than scenes with dense objects and very long videos.

\textbf{Performance on TempRMOT.} Table~\ref{tab:compar_spatial_ovis}--\ref{tab:compar_spatial_mot20} reports the spatial grounding and tracking results on three datasets. The model performs best on dataset OVIS, with a significantly higher HOTA score, indicating better overall tracking performance under occlusion. The AssA (overall association accuracy) scores are relatively high across all datasets, which suggests that the association component of the tracker can continuously associate with one object across frames after it has been detected. In contrast, the DetA scores of MOT17 and MOT20 are extremely low. This may be due to a combination of lots of unlabeled objects that meet the description being treated as false negatives and labeled objects that may not be detected correctly, suggesting that object detection remains the main bottleneck in dense and long-duration videos. These results highlight the importance of improving detection robustness and the annotation density to advance referring multi-object tracking in complex scenarios.

\textbf{Performance on FlashVTG.}
Table~\ref{tab:performance_flashvtg} presents evaluation results under two conditions: with and without non-maximum suppression (NMS, threshold 0.7). 
The model consistently achieves the best results on OVIS compared to MOT17 and MOT20 across all metrics, confirming the model's robustness in short videos, despite being occluded. Temporal localization is strongly correlated with video length. MOT videos average hundreds to thousands of frames, and performance degrades significantly. Applying NMS slightly improves R@5/10 and mAP for OVIS and MOT17 by removing redundant overlapping predictions, but it has no impact on R@1, mIoU, or MOT20. This is because R@1 and mIoU are determined solely by the best prediction, while the relatively low baseline detection quality in MOT20 further limits potential gains. These findings highlight that detection quality and redundancy management are critical for temporal action grounding performance.
\begin{table*}[]
	\centering
	\renewcommand{\arraystretch}{1.1}
	\scalebox{0.68}{
		\begin{tabular}{lccccccc}
			\specialrule{1.5pt}{0pt}{0pt}
            & \multicolumn{3}{c}{R1} & \multicolumn{3}{c}{mAP} &  \\ 
\multirow{-2}{*}{Methods} & @0.1 & @0.3 & @0.5 & @0.1 & @0.3 & @0.5 & \multirow{-2}{*}{mIoU} \\\hline\hline
		LD-DETR~\cite{zhao2025ld} & \underline{78.27} & 52.33 & \textbf{35.25} & \underline{84.16} & \underline{63.31} & \textbf{46.73} & \textbf{40.83} \\
       $\rm R^2$-Tuning~\cite{liu2024r} & 75.28 & \textbf{54.10} & 34.92 & 80.66 & \textbf{63.54} & \underline{46.07} & 38.50 \\
    VTimeLLM~\cite{huang2024vtimellm} & 64.30 & 42.24 & 20.84 & 61.76 & 40.50 & 19.84 & 27.73 \\
    d2vlm~\cite{zeng2025factorized}  & 69.40 & 51.00 & 21.62 & 69.95 & 51.72 & 21.58 & 29.30 \\
      TimeSuite~\cite{zeng2024timesuite}  & 47.23 & 22.62 & 12.31 & 45.21 & 21.54 & 11.57 & 18.21 \\ \hline
        EgoMask~\cite{liang2025finegrained} & 32.82 & 18.74 & 11.97 & 39.01 & 21.95 & 13.50 & 14.78 \\      
         Qwen2.5-VL~\cite{qwen2.5-VL} & 53.88 & 30.16 & 16.30 & 51.91 & 28.98 & 15.66 & 22.30 \\
        Qwen3-VL~\cite{qwen3technicalreport} & 2.44 & 0.11 & 0.00  & 2.40 & 0.11 & 0.00 & 1.19  \\
      InternVL3.5~\cite{wang2025internvl3_5} & 41.02 & 21.18 & 9.42  & 43.85 & 24.17 & 11.21 & 14.94  \\
      GLM-4.6V~\cite{vteam2025glm45vglm41vthinkingversatilemultimodal} & 51.73 & 30.12 & 16.05 & 51.56 & 29.58 & 15.61 & 20.76  \\\hline
    GPT-4.1 mini~\cite{openai_gpt41mini_2025}  & 62.54 & 41.07 & 27.30 & 62.88 & 43.28 & 27.48 & 30.80 \\
    GPT-5.4~\cite{openai_gpt54_2026} & 62.47 & 48.76 & \underline{35.17} & 65.94 & 52.03 & 38.60 & 35.55 \\
    % Claude-Sonnet-4.5~\cite{anthropic2025claudesonnet45} & & 47.70 & 30.80 & & & & 36.50 \\
    % Claude-Opus-4.6~\cite{anthropic2026claudeopus46} & & 38.43 & 27.45 & & & & 29.07 \\
    % Amazon Nova Lite~\cite{Intelligence2024} & & 42.19 & 28.12 & & & & 34.48 \\
    \hline
    \textbf{Ours} & \textbf{82.37} & \underline{52.66} & 33.37 & \textbf{85.63} & 61.22 & 43.98 & \underline{39.27}
\\
    \specialrule{1.5pt}{0pt}{0pt}
	\end{tabular}}
	\caption{Comparison with existing state-of-the-art methods on OVIS test set. \textbf{Bold} and \underline{underline}
    indicate the best and second-best results per column.}
	\label{tab:compar_temp_ovis}
	\vspace{-2mm}
\end{table*}

\begin{table*}[]
	\centering
	\renewcommand{\arraystretch}{1.1}
	\scalebox{0.68}{
    \begin{tabular}{lccccccc}
			\specialrule{1.5pt}{0pt}{0pt}
            & \multicolumn{3}{c}{R1} & \multicolumn{3}{c}{mAP} &  \\ 
\multirow{-2}{*}{Methods} & @0.1 & @0.3 & @0.5 & @0.1 & @0.3 & @0.5 & \multirow{-2}{*}{mIoU} \\\hline\hline
LD-DETR~\cite{zhao2025ld} & 27.25 & 12.22 & 4.21  & 26.59 & 11.93 & 4.69 & 8.90 \\
 $\rm R^2$-Tuning~\cite{liu2024r} & \underline{31.86} & \underline{15.83} & \underline{8.42} & \underline{35.69} & \underline{19.89} & \underline{11.17} & \underline{11.60} \\
    VTimeLLM~\cite{huang2024vtimellm} & \textbf{32.46} & 12.02 & 3.21 & 29.34 & 10.59 & 2.91 & 9.31 \\
    d2vlm~\cite{zeng2025factorized}  & 22.24 & 7.82 & 4.41 & 19.74 & 7.13 & 4.28 & 7.67  \\
     TimeSuite~\cite{zeng2024timesuite}  & 3.41 & 1.40 & 1.00 & 2.51 & 1.40 & 1.00 & 1.60 \\        \hline 
     EgoMask~\cite{liang2025finegrained} & 9.82 & 5.21 & 3.01 & 11.85 & 6.00 & 3.22 & 4.17 \\
         Qwen2.5-VL~\cite{qwen2.5-VL} & 9.22 &3.81 & 0.40 & 7.29 & 3.29 & 0.74 & 2.91  \\
         Qwen3-VL~\cite{qwen3technicalreport} & 1.20 & 0.20 & 0.00  & 1.20 & 0.20 & 0.00 & 0.57  \\
      InternVL3.5~\cite{wang2025internvl3_5} & 14.63 & 6.61 & 3.21 & 14.44 & 6.09 & 3.40 & 5.49  \\
       GLM-4.6V~\cite{vteam2025glm45vglm41vthinkingversatilemultimodal} & 16.36 & 6.54 & 2.57 & 14.44 & 6.90 & 3.19 & 5.52  \\\hline
   GPT-4.1 mini~\cite{openai_gpt41mini_2025} & 21.87 & 11.68 & 6.58 & 17.10 & 8.86 & 5.32 & 9.36 \\
    GPT-5.4~\cite{openai_gpt54_2026} & 29.86 & \textbf{20.64} & \textbf{13.63} & 27.70 & 19.27 & \textbf{13.11} & \textbf{15.46}\\
        % Claude-Sonnet-4.5~\cite{anthropic2025claudesonnet45} & & 20.40 & 14.50 & & & & 17.00 \\
    % Claude-Opus-4.6~\cite{anthropic2026claudeopus46} & & 9.54 & 6.07 & & & & 7.96 \\
        % Amazon Nova Lite~\cite{Intelligence2024} & & 11.58 & 8.87 & & & & 10.17 \
        hline
    \textbf{Ours} & 30.86 & 13.23 & 6.41 & \textbf{43.67} & \textbf{21.79} & 8.97 & 10.48\\

    \specialrule{1.5pt}{0pt}{0pt}
	\end{tabular}}
	\caption{Comparison with existing state-of-the-art methods on MOT17 test set. \textbf{Bold} and \underline{underline}
    indicate the best and second-best results per column.}
	\label{tab:compar_temp_mot17}
	\vspace{-2mm}
\end{table*}

\begin{table*}[]
	\centering
	\renewcommand{\arraystretch}{1.1}
	\scalebox{0.68}{
		\begin{tabular}{lccccccc}
			\specialrule{1.5pt}{0pt}{0pt}
            & \multicolumn{3}{c}{R1} & \multicolumn{3}{c}{mAP} &  \\ 
\multirow{-2}{*}{Methods} & @0.1 & @0.3 & @0.5 & @0.1 & @0.3 & @0.5 & \multirow{-2}{*}{mIoU} \\\hline\hline
LD-DETR~\cite{zhao2025ld} & 0.93 & 0.69 & 0.69 & 0.97 & 0.42 & 0.29 & 0.55 \\
 $\rm R^2$-Tuning~\cite{liu2024r} & 18.29 & \underline{9.72} & \textbf{6.02} & \underline{17.67} & \textbf{10.29} & \underline{4.54} & \underline{7.33} \\	
      VTimeLLM~\cite{huang2024vtimellm} & \textbf{21.06} &  6.25 & 0.93 & 9.56 & 2.41 & 0.69 & 6.17 \\
    d2vlm~\cite{zeng2025factorized} & 0.00 & 0.00 & 0.00 & 0.00 & 0.00 & 0.00 & 0.00
          \\     TimeSuite~\cite{zeng2024timesuite} & 1.16 & 0.00 & 0.00 & 0.58 & 0.00 & 0.00 & 0.15 \\ \hline
           EgoMask~\cite{liang2025finegrained} & 0.93 & 0.46 & 0.00 & 2.88 & 0.90 & 0.22 & 0.42 \\
            Qwen2.5-VL~\cite{qwen2.5-VL} & 2.55 & 2.08 & 1.16 & 1.74& 1.01& 0.46 & 1.53 \\
               Qwen3-VL~\cite{qwen3technicalreport} & 1.62 & 0.46 & 0.00 & 0.97 & 0.30 & 0.00 & 0.39  \\
            InternVL3.5~\cite{wang2025internvl3_5}  & 7.41 & 2.78 & 0.69 & 6.84 & 3.04 & 0.71 & 2.25  \\
    GLM-4.6V~\cite{vteam2025glm45vglm41vthinkingversatilemultimodal} & 5.61 & 1.28 &0.26 & 4.33 & 1.19 &0.29 & 1.32  \\ \hline 
     GPT-4.1 mini~\cite{openai_gpt41mini_2025}  & 7.11 & 2.21 & 0.74 & 5.59 & 1.80 & 0.66 & 2.36 \\
    GPT-5.4~\cite{openai_gpt54_2026} & 5.79 & 1.62 & 0.46 & 7.07 & 1.42 & 0.56 & 1.92 \\
      % Claude-Sonnet-4.5~\cite{anthropic2025claudesonnet45} & & 2.90 & 0.70 & & & & 3.80 \\
    % Claude-Opus-4.6~\cite{anthropic2026claudeopus46} & & 2.74 & 0.27 & & & & 3.08     \\
    % Amazon Nova Lite~\cite{Intelligence2024} & & 0.00 & 0.00 & & & & 0.31 \\\hline
        \hline
\textbf{Ours} & \underline{20.14} & \textbf{9.95} & \underline{4.17} & \textbf{20.56} & \underline{10.07} & \textbf{5.47} & \textbf{7.62}

      \\\specialrule{1.5pt}{0pt}{0pt}
	\end{tabular}}
	\caption{Comparison with existing state-of-the-art methods on MOT20 test set. \textbf{Bold} and \underline{underline}
    indicate the best and second-best results per column.}
	\label{tab:compar_temp_mot20}
	\vspace{-2mm}
\end{table*}

\begin{table}[]
	\centering
	\renewcommand{\arraystretch}{1.1}
	\scalebox{0.88}{
		\begin{tabular}{lcccccccc}
			\specialrule{1.5pt}{0pt}{0pt}
			%\rowcolor{mygray} 
			Methods & HOTA & DetA & AssA & DetRe & DetPr & AssRe & AssPr & LocA \\ \hline\hline
       
TransRMOT~\cite{wu2023referring} & 2.834 & 1.173 & 9.462 & 1.448 & 4.718 & 11.541 & 54.445 & 59.350 \\
     DKGTrack~\cite{li2025language} & \underline{17.847} & \underline{8.669} & 38.013 & 15.835 & \underline{15.311} & 51.030 & \underline{55.515} & 74.457 \\
  Sa2VA~\cite{yuan2025sa2va} & 7.759 & 1.493 & \underline{41.354}  & \textbf{35.875} & 1.531 & \textbf{64.448} & 46.637 & \textbf{83.189} \\\hline
      EgoMask~\cite{liang2025finegrained} & 12.086 & 4.267 & 36.909 & 8.278 & 7.768 & \underline{59.121} & 44.465 & 71.893 \\
      Qwen2.5-VL~\cite{qwen2.5-VL} & 5.199 & 1.282 & 25.163 & 2.005 & 3.306 & 36.424 & 54.156 & 64.218  \\ 
   Qwen3-VL~\cite{qwen3technicalreport} 
& 2.463 & 0.599 & 11.033 & 0.860 & 1.862 & 24.188 & 26.191 & 61.279 \\
InternVL3.5~\cite{wang2025internvl3_5} & 0.988 & 0.182 & 6.657 & 0.235 & 0.786 & 8.389 & 32.774 & 63.371 \\
GLM-4.6V~\cite{vteam2025glm45vglm41vthinkingversatilemultimodal} 
& 0.728 & 0.087 & 7.083 & 0.163 & 0.183 & 10.274 & 11.613 & 69.633 \\      \hline
GPT-4.1 mini~\cite{openai_gpt41mini_2025} & 4.743 & 1.077 & 25.049 & 2.131 & 2.046 & 34.794 & 36.410 & 60.669 \\
GPT-5.4~\cite{openai_gpt54_2026}  & 11.622 & 4.461 & 39.892 & 10.451 & 6.590 & 56.869 & 47.758 & 62.268 \\
    % Claude-Sonnet-4.5~\cite{anthropic2025claudesonnet45} &&&&&&& \\
Claude-Opus-4.6~\cite{anthropic2026claudeopus46} & 9.352 & 4.005 & 25.972 & 9.049 & 6.045 & 40.598 & 33.925 & 60.344 \\
\hline
         \textbf{Ours} & \textbf{22.734} & \textbf{11.234} & \textbf{46.625} & \underline{20.020} & \textbf{19.910} & 56.555 & \textbf{67.611} & \underline{82.768} \\\hline

            \specialrule{1.5pt}{0pt}{0pt}
	       \end{tabular}}
	\caption{Comparison with existing state-of-the-art methods on OVIS test set. \textbf{Bold} and \underline{underline}
    indicate the best and second-best results per column.}
	\label{tab:compar_spatial_ovis}
	\vspace{-3mm}
\end{table}

\begin{table}[!t]
	\centering
	\renewcommand{\arraystretch}{1.1}
	\scalebox{0.88}{
		\begin{tabular}{lcccccccc}
			\specialrule{1.5pt}{0pt}{0pt}
			%\rowcolor{mygray} 
			Dataset & HOTA & DetA & AssA & DetRe & DetPr & AssRe & AssPr & LocA \\ \hline\hline
		TransRMOT~\cite{wu2023referring} & 0.127 & 0.011 &  1.606 & 0.015 & 0.036 & 1.656 & 33.080 & 68.130 \\
     DKGTrack~\cite{li2025language} & \textbf{1.585} & \textbf{0.177} & \textbf{15.507} & \underline{0.245} & \textbf{0.630}& \underline{23.550} & \textbf{62.471} & 78.455 \\ 
  Sa2VA~\cite{yuan2025sa2va} & 0.450 & 0.017 & \underline{14.149} & \textbf{0.344} & 0.018 & \textbf{35.858} & 18.846 & 71.799 \\\hline
  EgoMask~\cite{liang2025finegrained} & 0.028 & 0.005 & 0.208 & 0.005 & 0.046 & 1.202 & 0.251 & 78.823 \\
   Qwen2.5-VL~\cite{qwen2.5-VL} & 0.030 & 0.002 & 0.456 & 0.003 & 0.019 & 0.701 & 1.139 & \underline{90.531} \\
   Qwen3-VL~\cite{qwen3technicalreport} 
& 0.088 & 0.005 & 1.891 & 0.005 & 0.148 & 1.905 & 29.268 & 73.656 \\

InternVL3.5~\cite{wang2025internvl3_5} 
& 0.014 & 0.002 & 0.167 & 0.002 & 0.054 & 0.172 & 8.192 & 66.098 \\

GLM-4.6V~\cite{vteam2025glm45vglm41vthinkingversatilemultimodal} 
& 0.002 & 0.001 & 0.031 & 0.001 & 0.001 & 0.041 & 0.124 & \textbf{95.051} \\    \hline 

    GPT-4.1 mini~\cite{openai_gpt41mini_2025}  & 0.004 & 0.001 & 0.088 & 0.001 & 0.006 & 0.107 & 2.106 & 82.658 \\
    GPT-5.4~\cite{openai_gpt54_2026}  & 0.086 & 0.013 & 0.669 & 0.014 & 0.112 & 1.149 & 2.520 & 63.337 \\
    % Claude-Sonnet-4.5~\cite{anthropic2025claudesonnet45} &&&&&&& \\
      Claude-Opus-4.6~\cite{anthropic2026claudeopus46} & 0.054 & 0.006 & 0.523 & 0.007 & 0.119 & 0.795 & 7.896 & 67.867 \\
    \hline
        \textbf{Ours} & \underline{0.597} & \underline{0.043} & 9.172 & 0.049 & \underline{0.366} & 11.659 & \underline{56.583} & 75.968 \\

\specialrule{1.5pt}{0pt}{0pt}
	       \end{tabular}}
	\caption{Comparison with existing state-of-the-art methods on MOT17 test set. \textbf{Bold} and \underline{underline}
    indicate the best and second-best results per column.}
	\label{tab:compar_spatial_mot17}
	\vspace{-3mm}
\end{table}

\begin{table}[]
	\centering
	\renewcommand{\arraystretch}{1.1}
	\scalebox{0.88}{
		\begin{tabular}{lcccccccc}
			\specialrule{1.5pt}{0pt}{0pt}
			%\rowcolor{mygray} 
			Dataset & HOTA & DetA & AssA & DetRe & DetPr & AssRe & AssPr & LocA \\ \hline\hline
TransRMOT~\cite{wu2023referring} & \textbf{0.430} & \textbf{0.090} & \underline{2.362} & \textbf{0.575} & \underline{0.106} & 2.572 & \textbf{42.891} & 57.795 \\ 
     DKGTrack~\cite{li2025language} & 0.063 & 0.008 & 0.559 &  0.010 & 0.030 & 0.878 & \underline{39.315} & 65.412 \\
  Sa2VA~\cite{yuan2025sa2va} & 0.149 & 0.019 & 1.257 & 0.031 & 0.051 & \textbf{14.983} & 1.616 & 62.585\\\hline
  EgoMask~\cite{liang2025finegrained} & 0.000 & 0.000 & 0.000 & 0.000 & 0.000 & 0.000 & 0.000 & \textbf{100.000}  \\
  Qwen2.5-VL~\cite{qwen2.5-VL} & 0.000 & 0.000 & 0.000 & 0.000 & 0.000 & 0.000 & 0.000 & \textbf{100.000}  \\
Qwen3-VL~\cite{qwen3technicalreport} 
& 0.021 & 0.001 & 0.874 & 0.001 & 0.034 & 0.934 & 12.662 & 67.630 \\
InternVL3.5~\cite{wang2025internvl3_5} 
& 0.006 & 0.001 & 0.278 & 0.001 & 0.012 & 0.560 & 2.544 & 83.248 \\
        GLM-4.6V~\cite{vteam2025glm45vglm41vthinkingversatilemultimodal} & 0.000 & 0.000 & 0.000 & 0.000 & 0.000 & 0.000 & 0.000 & \textbf{100.000}  \\    \hline 
    GPT-4.1 mini~\cite{openai_gpt41mini_2025}  & 0.000 & 0.000 & 0.000 & 0.000 & 0.000 & 0.000 & 0.000 & \textbf{100.000}  \\
      GPT-5.4~\cite{openai_gpt54_2026}  & 0.020 & 0.002 & 0.331 & 0.002 & 0.052 & 0.530 & 0.872 & 82.690 \\
    % Claude-Sonnet-4.5~\cite{anthropic2025claudesonnet45} &&&&&&& \\
      Claude-Opus-4.6~\cite{anthropic2026claudeopus46} & 0.035 & 0.003 & 0.690 & 0.003 & 0.136 & 0.707 & 23.825 & 63.397 \\
    \hline
\textbf{Ours} & \underline{0.429} & \underline{0.066} & \textbf{2.971} & \underline{0.096} & \textbf{0.206} & \underline{5.539} & 13.112 & 64.431 \\  \hline

  \specialrule{1.5pt}{0pt}{0pt}
	       \end{tabular}}
	\caption{Comparison with existing state-of-the-art methods on MOT20 test set. \textbf{Bold} and \underline{underline}
    indicate the best and second-best results per column.}
	\label{tab:compar_spatial_mot20}
	\vspace{-3mm}
\end{table}

\begin{table}[]
	\centering
	\renewcommand{\arraystretch}{1.1}
	\scalebox{0.72}{
		\begin{tabular}{lccccccccccccc}
			\specialrule{1.5pt}{0pt}{0pt}
			%\rowcolor{mygray} 
			% \cellcolor{mygray} & \multicolumn{3}{c}{ \cellcolor{mygray} R1} & \multicolumn{3}{c}{ \cellcolor{mygray} R5} & \multicolumn{3}{c}{ \cellcolor{mygray} R10} & \multicolumn{3}{c}{ \cellcolor{mygray} mAP} & \cellcolor{mygray}  \\ 
			% \rowcolor{mygray} 
			% \multirow{-2}{*}{\cellcolor{mygray} Dataset} & @0.1 & @0.3 & 0.5 & @0.1 & @0.3 & 0.5 & @0.1 & @0.3 & 0.5 & @0.1 & @0.3 & 0.5 & \multirow{-2}{*}{\cellcolor{mygray} mIoU} \\

            & \multicolumn{3}{c}{R1} & \multicolumn{3}{c}{R5} & \multicolumn{3}{c}{R10} & \multicolumn{3}{c}{mAP} &  \\ 
\multirow{-2}{*}{Dataset} & @0.1 & @0.3 & @0.5 & @0.1 & @0.3 & @0.5 & @0.1 & @0.3 & @0.5 & @0.1 & @0.3 & @0.5 & \multirow{-2}{*}{mIoU} \\

			\hline
			\hline
			OVIS & 82.37 & 52.66 & 33.37 & 90.91 & 72.17 & 56.87 & 94.68 & 83.59 & 71.51 & 85.25 & 59.98 & 42.12 & 39.27 \\
OVIS\textsuperscript{\dag} & 82.37 & 52.66 & 33.37 & \textbf{91.57} & \textbf{75.06} & \textbf{59.31} & \textbf{94.9} & \textbf{85.03} & \textbf{75.06} & \textbf{85.63} & \textbf{61.22} & \textbf{43.98} & 39.27 \\\hline 
MOT17 & 30.86 & 13.23 & 6.41 & 69.74 & 38.08 & 14.03 & 79.76 & 48.5 & 21.04 & 42.45 & 21.78 & 8.69 & 10.48 \\
MOT17\textsuperscript{\dag} & 30.86 & 13.23 & 6.41 & \textbf{74.15} & \textbf{39.48} & \textbf{15.63} & \textbf{83.57} & \textbf{48.9} & \textbf{21.24} & \textbf{43.67} & \textbf{21.79} & \textbf{8.97} & 10.48 \\\hline 
MOT20 & 20.14 & 9.95 & 4.17 & 48.15 & 22.45 & 12.73 & 61.81 & 31.02 & 20.83 & 20.56 & 10.07 & 5.47 & 7.62 \\
MOT20\textsuperscript{\dag} & 20.14 & 9.95 & 4.17 & 48.15 & 22.45 & 12.73 & 61.81 & 31.02 & 20.83 & 20.56 & 10.07 & 5.47 & 7.62 \\\specialrule{1.5pt}{0pt}{0pt}
	\end{tabular}}
	\caption{Performance on the different datasets using FlashVTG \cite{Cao_2025_WACV}. Unlabeled datasets do not use NMS. Datasets marked with \dag\ use NMS 0.7. The higher score is highlighted in bold. Applying NMS will slightly improve R@5/10 and mAP for OVIS and MOT17, with no impact on MOT20.}
	\label{tab:performance_flashvtg}
	\vspace{-2mm}
\end{table}

\subsection{Results Analysis}

\begin{figure}[]
  \centering
  % 左图
  \begin{subfigure}[b]{0.48\textwidth}
    \centering
    \includegraphics[width=\linewidth]{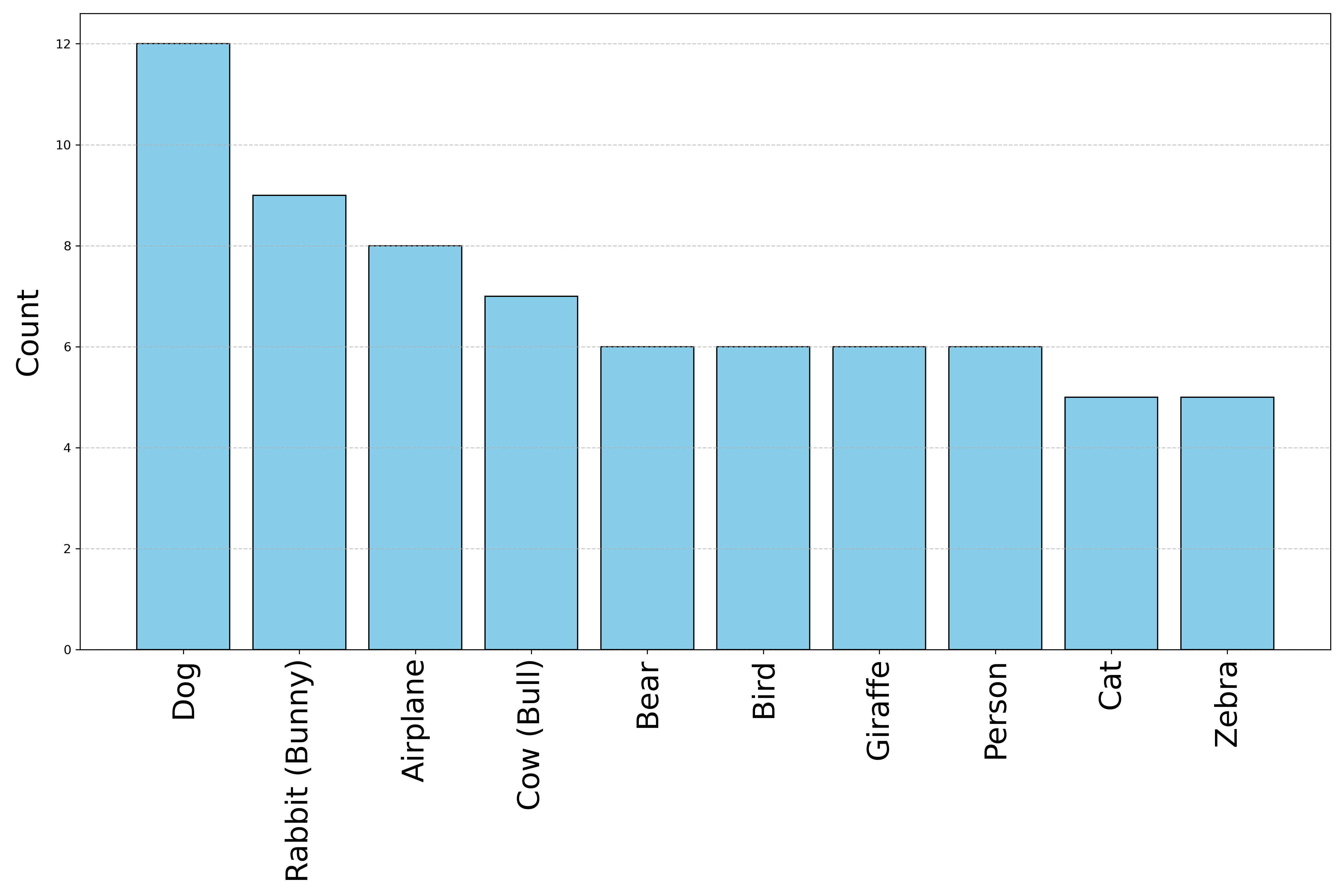}
    \caption{Top 10 categories of referent objects}
    \label{fig:categories_over}
  \end{subfigure}\hfill
  % 右图
  \begin{subfigure}[b]{0.48\textwidth}
    \centering
    \includegraphics[width=\linewidth]{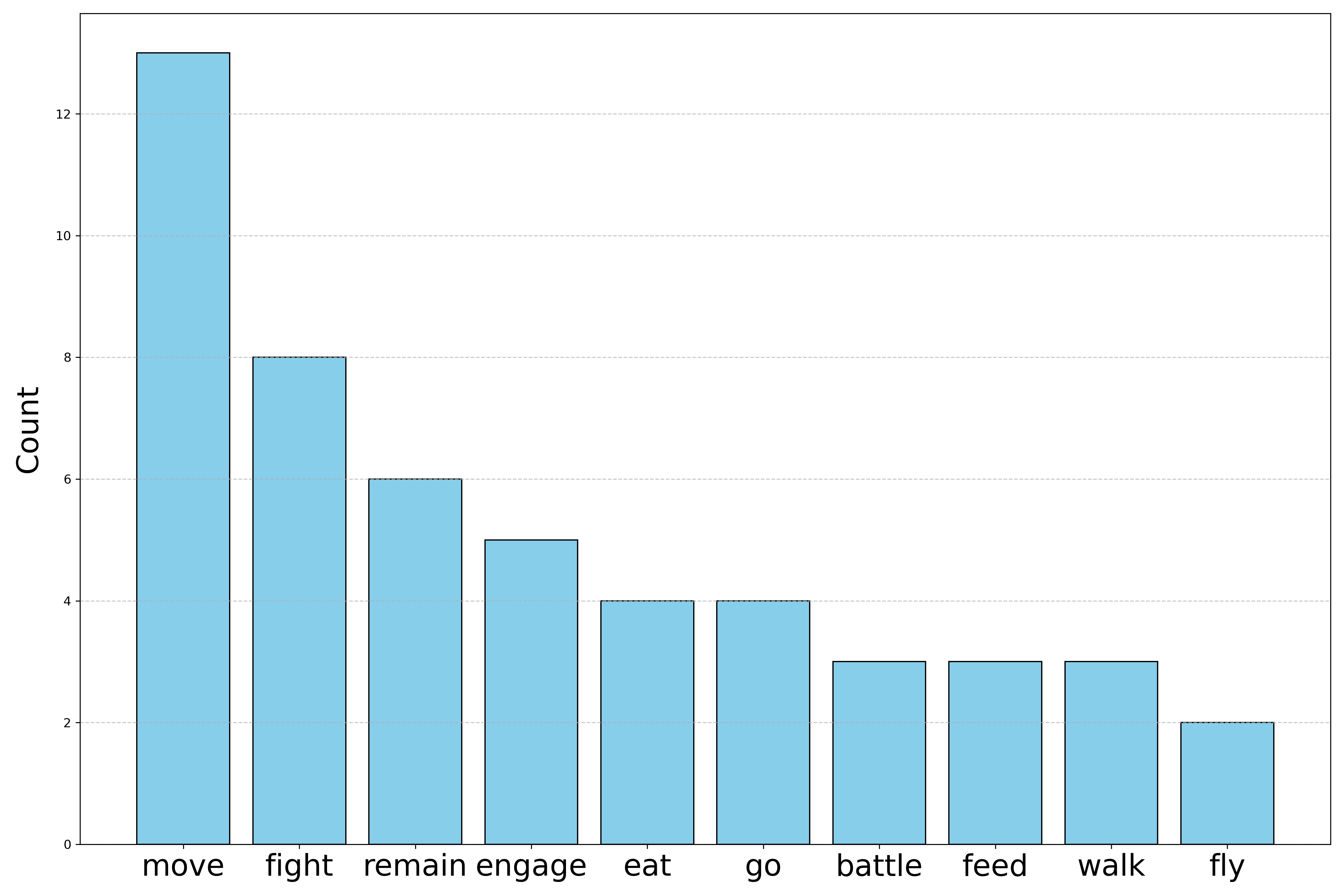}
    \caption{Top 10 actions of referent objects}
    \label{fig:actions_over}
  \end{subfigure}

  \caption{\textbf{Statistics on OVIS dataset}. High performance on categories like Dog, Rabbit, and Airplane. Dynamic actions like move, fight lead successful sequences.
\vspace{-3mm}}
  \label{fig:ovis_svag}
\end{figure}

To better understand the performance of TempRMOT, we analyze results on the OVIS dataset. We focus on sequences with high performance and identify the top 10 referent categories and actions, as shown in Fig.~\ref{fig:ovis_svag}.

The high-performing cases are dominated by animal categories (e.g., dog, rabbit). Conversely, some frequent categories in the dataset (e.g., fish, poultry) are underrepresented, suggesting that action observability and motion diversity are primary drivers of effective grounding. 

The distribution of verbs further supports this finding: dynamic actions (e.g., ``move'', ``fight'', ``eat'') dominate successful sequences, though static states (e.g., ``remain'') also appear. This indicates that both explicit motion and temporal continuity are leveraged by the model, aligning with the task objective of spatiotemporal grounding based on action queries.

\section{Supplementary: Prompt Configurations and Ablation Studies}

\subsection{Prompt Designs}
\label{sec:prompts}

We evaluate LVLMs using four structured prompts: one for spatial
grounding (Prompt~A.1) and three for temporal grounding
(Prompts~A.2--A.4), listed below. Prompt~A.2 targets single-instance
temporal grounding; Prompt~A.3 extends this to multi-instance
scenarios; Prompt~A.4 further conditions on provided bounding boxes
to isolate temporal reasoning from spatial detection.

\begin{neuprompt}
{A.1 Prompt for Spatial Grounding}
\label{prompt:spatial_grounding}
Given the query \{query\}, for each frame, detect and localize all the visual contents described by the given textual query in JSON format.\newline
If the visual content does not exist in a frame, skip that frame.\newline\newline
Output Format:\newline
[\{\{\newline
\hspace*{2em}"object\_id": 1,\newline
\hspace*{2em}"frames": [\newline
\hspace*{4em}\{\{"t": 1, "bbox": [x\_min, y\_min, x\_max, y\_max]\}\},\newline
\hspace*{4em}\{\{"t": 2, "bbox": [x\_min, y\_min, x\_max, y\_max]\}\},\newline
\hspace*{2em}]\newline
\}\},\newline
\{\{\newline
\hspace*{2em}"object\_id": 2,\newline
\hspace*{2em}"frames": [\newline
\hspace*{4em}\{\{"t": 2, "bbox": [x\_min, y\_min, x\_max, y\_max]\}\},\newline
\hspace*{4em}\{\{"t": 4, "bbox": [x\_min, y\_min, x\_max, y\_max]\}\},\newline
\hspace*{2em}]\newline
\}\}]\newline\newline
Notes:\newline
- Do NOT include explanations.\newline
- Only output the JSON object described above.
\end{neuprompt}

\begin{neuprompt}
{A.2 Prompt for General Temporal Grounding}
\label{prompt:temporal_grounding_1}
You are performing temporal grounding for a video.\newline\newline
Event: "\{query\}"\newline\newline
Please answer using EXACTLY the following format:\newline
\{query\} from frame <start\_frame> to <end\_frame>\newline\newline
Only output the final answer in ONE line, no explanation.
\end{neuprompt}

\begin{neuprompt}
{A.3 Prompt for Multi-Instance Temporal Grounding}
\label{prompt:temporal_grounding_2}
You are performing temporal grounding for a video.\newline\newline
Event: "\{query\}"\newline\newline
There may be multiple objects that satisfy the query.\newline\newline
For each object:\newline
- Identify its object\_id\newline
- Identify its start frame\newline
- Identify its end frame\newline\newline
Please answer using EXACTLY the following format:\newline
object\_id: <id>\newline
start\_frame: <time>\newline
end\_frame: <time>\newline\newline
One block per object. No explanation.
\end{neuprompt}

\begin{neuprompt}
{A.4 Prompt for Multi-Instance Temporal Grounding with Bounding Boxes}
\label{prompt:temporal_grounding_3}
You are checking whether an event occurs for multiple objects in the video.\newline\newline
Event: "\{query\}"\newline\newline
Bounding boxes for all objects are provided for each frame.\newline\newline
For each object, output EXACTLY one line:\newline\newline
If the event happens:\newline
object <obj\_id>: Yes, (start\_frame, end\_frame, confidence\_score)\newline
or if it does NOT happen:\newline
object <obj\_id>: No, it does not happen.\newline\newline
IMPORTANT RULES:\newline
- Output results ONLY for the object IDs that appear in the bounding boxes.\newline
- Do NOT create additional object IDs (do NOT invent object 0 or new objects).\newline
- You MUST output one line for EVERY object with bounding boxes.\newline
- Do NOT skip objects.\newline
- Each object has AT MOST one (start, end, score).\newline
- No explanation.
\end{neuprompt}

\subsection{LVLM Configuration Details}
\label{sec:lvlm_configs}

Table~\ref{tab:lvlm_configs} summarizes the prompt and input
configuration used for each LVLM across spatial and temporal
grounding evaluations. \textit{Resize} denotes aspect-ratio-preserving
resizing with the longer side capped at 448 pixels and the shorter
side ensured to be at least 252 pixels. Fixed values (32, 128, 256)
denote the number of uniformly sampled frames.

\begin{table}[!t]
    \centering
    \renewcommand{\arraystretch}{1.1}
    \resizebox{\columnwidth}{!}{
    \begin{tabular}{lcccc}
        \specialrule{1.5pt}{0pt}{0pt}
        \multirow{2}{*}{Model} 
          & \multicolumn{2}{c}{Temporal Grounding}
          & \multicolumn{2}{c}{Spatial Grounding} \\
        \cmidrule(lr){2-3} \cmidrule(lr){4-5}
          & OVIS & MOT17 / MOT20 & All Datasets & Input \\
        \hline\hline
        Qwen2.5-VL & A.2 + resize    & A.3 + 256 frames & A.1 & resize \\
        Qwen3-VL   & A.3 + 256 frames & A.3 + 256 frames & A.1 & 256 frames \\
        InternVL3.5 & A.3 + resize + 32f & A.3 + resize + 32f & A.1 & resize + 32f \\
        GLM-4.6V   & A.3 + resize + 128f & A.3 + resize + 128f & A.1 & resize + 256f \\
         GPT-4.1 mini  & A.3 + resize + 128f & A.3 + resize + 128f & A.1 & resize + 128f \\
        GPT-5.4  & A.3 + resize + 128f & A.3 + resize + 128f & A.1 & resize + 128f \\
          Claude-Sonnet-4.5  & A.3 + resize + 64f & A.3 + resize + 64f & &  \\
          Claude-Opus-4.6 & A.3 + resize + 96f & A.3 + resize + 96f & A.1 & resize + 96f \\

        \specialrule{1.5pt}{0pt}{0pt}
    \end{tabular}}
    \caption{Prompt and input configuration for each LVLM across
    spatial and temporal grounding evaluations.}
    \label{tab:lvlm_configs}
\end{table}

\subsection{LVLM Failure Modes}
\label{sec:failure_modes}

We observe three recurring failure modes when evaluating LVLMs on
SVAG. First, \textbf{frame index overflow}: models predict frame
indices beyond the actual video length. Second, \textbf{bounding
box overflow}: predicted coordinates exceed the frame dimensions.
Third, \textbf{static box hallucination}: the model outputs identical
bounding boxes across all frames, indicating a failure to track
motion over time. These failure modes are most prevalent on MOT20,
where dense crowds and long sequences amplify the model's tendency
to hallucinate structured outputs.

For Sa2VA~\cite{yuan2025sa2va}, predicted masks are converted to
bounding boxes, and overlapping instances are merged into a single
object. This introduces potential object-ID consistency issues across
frames. Sampling strides are applied to manage sequence length:
MOT17-03 and MOT20-03 use a stride of 2; MOT20-05 uses a stride of 3.

\subsection{Prompt Ablation Studies}
\label{sec:prompt_ablation}

Tables~\ref{tab:ablation_ovis_qwen2.5}--\ref{tab:ablation_mot20_qwen2.5}
report temporal grounding ablations across prompt variants and input
configurations for Qwen2.5-VL on all three datasets.
Tables~\ref{tab:ablation_spatial_ovis_qwen2.5}--\ref{tab:ablation_spatial_mot20_qwen2.5}
report the corresponding spatial grounding ablations.

On OVIS, Prompt~A.2 with resize achieves the best overall temporal
performance (mIoU: 22.30), suggesting that single-instance prompting
is more reliable on shorter, less crowded sequences. On MOT17 and
MOT20, Prompt~A.3 with 256 frames consistently outperforms resize
variants, indicating that fixed frame sampling better handles
long-form videos where dynamic resolution leads to excessive
token consumption. Spatial grounding performance collapses on
MOT20 under all configurations, consistent with results in the
main paper.

\begin{table*}[!t]
    \centering
    \renewcommand{\arraystretch}{1.1}
    \resizebox{\textwidth}{!}{
    \begin{tabular}{lccccccc}
        \specialrule{1.5pt}{0pt}{0pt}
        & \multicolumn{3}{c}{R1} & \multicolumn{3}{c}{mAP} & \\
        \multirow{-2}{*}{Configuration} & @0.1 & @0.3 & @0.5 & @0.1 & @0.3 & @0.5 & \multirow{-2}{*}{mIoU} \\
        \hline\hline
        Prompt A.2 + resize      & \textbf{53.88} & \textbf{30.16} & 16.30 & \textbf{51.91} & 28.98 & 15.66 & \textbf{22.30} \\
        Prompt A.4 + resize      & 37.25 & 26.39 & \textbf{16.30} & 35.80 & 25.41 & 15.70 & 18.56 \\
        Prompt A.3 + resize      & 45.89 & 28.22 & 19.22 & 43.61 & 26.57 & 18.47 & 21.89 \\
        Prompt A.3 + 256 frames  & 34.15 & 16.30 & 7.21  & 36.55 & \textbf{18.80} & \textbf{8.40}  & 12.18 \\
        \specialrule{1.5pt}{0pt}{0pt}
    \end{tabular}}
    \caption{Temporal grounding ablation on OVIS test set (Qwen2.5-VL).}
    \label{tab:ablation_ovis_qwen2.5}
\end{table*}

\begin{table*}[!t]
    \centering
    \renewcommand{\arraystretch}{1.1}
    \resizebox{\textwidth}{!}{
    \begin{tabular}{lccccccc}
        \specialrule{1.5pt}{0pt}{0pt}
        & \multicolumn{3}{c}{R1} & \multicolumn{3}{c}{mAP} & \\
        \multirow{-2}{*}{Configuration} & @0.1 & @0.3 & @0.5 & @0.1 & @0.3 & @0.5 & \multirow{-2}{*}{mIoU} \\
        \hline\hline
        Prompt A.2 + resize      & 4.21 & 1.00 & 0.40 & 4.21 & 1.00 & 0.40 & 1.40 \\
        Prompt A.4 + resize      & 1.80 & 0.80 & 0.60 & 1.94 & 0.85 & 0.60 & 0.66 \\
        Prompt A.3 + resize      & 3.82 & 0.60 & 0.40 & 3.84 & 0.55 & 0.32 & 1.12 \\
        Prompt A.3 + 256 frames  & \textbf{9.22} & \textbf{3.81} & \textbf{0.40} & \textbf{7.29} & \textbf{3.29} & \textbf{0.74} & \textbf{2.91} \\
        \specialrule{1.5pt}{0pt}{0pt}
    \end{tabular}}
    \caption{Temporal grounding ablation on MOT17 test set (Qwen2.5-VL).}
    \label{tab:ablation_mot17_qwen2.5}
\end{table*}

\begin{table*}[!t]
    \centering
    \renewcommand{\arraystretch}{1.1}
    \resizebox{\textwidth}{!}{
    \begin{tabular}{lccccccc}
        \specialrule{1.5pt}{0pt}{0pt}
        & \multicolumn{3}{c}{R1} & \multicolumn{3}{c}{mAP} & \\
        \multirow{-2}{*}{Configuration} & @0.1 & @0.3 & @0.5 & @0.1 & @0.3 & @0.5 & \multirow{-2}{*}{mIoU} \\
        \hline\hline
        Prompt A.2 + resize      & 0.00 & 0.00 & 0.00 & 0.00 & 0.00 & 0.00 & 0.03 \\
        Prompt A.4 + resize      & 0.23 & 0.00 & 0.00 & 0.08 & 0.00 & 0.00 & 0.15 \\
        Prompt A.3 + resize      & 1.18 & 0.47 & 0.47 & 0.47 & 0.17 & 0.17 & 0.43 \\
        Prompt A.3 + 256 frames  & \textbf{2.55} & \textbf{2.08} & \textbf{1.16} & \textbf{1.74} & \textbf{1.01} & \textbf{0.46} & \textbf{1.53} \\
        \specialrule{1.5pt}{0pt}{0pt}
    \end{tabular}}
    \caption{Temporal grounding ablation on MOT20 test set (Qwen2.5-VL).}
    \label{tab:ablation_mot20_qwen2.5}
\end{table*}

\begin{table}[!t]
    \centering
    \renewcommand{\arraystretch}{1.1}
    \resizebox{\columnwidth}{!}{
    \begin{tabular}{lcccccccc}
        \specialrule{1.5pt}{0pt}{0pt}
        Configuration & HOTA & DetA & AssA & DetRe & DetPr & AssRe & AssPr & LocA \\
        \hline\hline
        Prompt A.1 + resize       & \textbf{5.199} & \textbf{1.282} & \textbf{25.163} & \textbf{2.005} & \textbf{3.306} & \textbf{36.424} & \textbf{54.156} & 64.218 \\
        Prompt A.1 + 256 frames   & 1.539 & 0.329 & 7.903 & 0.358 & 3.782 & 12.751 & 51.583 & \textbf{62.053} \\
        \specialrule{1.5pt}{0pt}{0pt}
    \end{tabular}}
    \caption{Spatial grounding ablation on OVIS test set (Qwen2.5-VL).}
    \label{tab:ablation_spatial_ovis_qwen2.5}
\end{table}

\begin{table}[!t]
    \centering
    \renewcommand{\arraystretch}{1.1}
    \resizebox{\columnwidth}{!}{
    \begin{tabular}{lcccccccc}
        \specialrule{1.5pt}{0pt}{0pt}
        Configuration & HOTA & DetA & AssA & DetRe & DetPr & AssRe & AssPr & LocA \\
        \hline\hline
        Prompt A.1 + resize       & \textbf{0.030} & \textbf{0.002} & \textbf{0.456} & \textbf{0.003} & \textbf{0.019} & \textbf{0.701} & 1.139  & 90.531 \\
        Prompt A.1 + 256 frames   & 0.004 & 0.001 & 0.180 & 0.001 & 0.029 & 0.180 & \textbf{15.789} & \textbf{87.322} \\
        \specialrule{1.5pt}{0pt}{0pt}
    \end{tabular}}
    \caption{Spatial grounding ablation on MOT17 test set (Qwen2.5-VL).}
    \label{tab:ablation_spatial_mot17_qwen2.5}
\end{table}

\begin{table}[!t]
    \centering
    \renewcommand{\arraystretch}{1.1}
    \resizebox{\columnwidth}{!}{
    \begin{tabular}{lcccccccc}
        \specialrule{1.5pt}{0pt}{0pt}
        Configuration & HOTA & DetA & AssA & DetRe & DetPr & AssRe & AssPr & LocA \\
        \hline\hline
        Prompt A.1 + resize       & 0.000 & 0.000 & 0.000 & 0.000 & 0.000 & 0.000 & 0.000 & 100.000 \\
        Prompt A.1 + 256 frames   & 0.000 & 0.000 & 0.000 & 0.000 & 0.000 & 0.000 & 0.000 & 100.000 \\
        \specialrule{1.5pt}{0pt}{0pt}
    \end{tabular}}
    \caption{Spatial grounding ablation on MOT20 test set (Qwen2.5-VL).
    All configurations score zero, confirming that MOT20 poses an
    intractable challenge for current LVLMs under spatial grounding.}
    \label{tab:ablation_spatial_mot20_qwen2.5}
\end{table}

\section{Competition}
\label{sec:competition}
We organize an ICCV 2025 Workshop\footnote{\url{https://motchallenge.net/workshops/bmtt2025/}} dedicated to the SVAG task. At the 8th edition of the BMTT workshop, the focus is on action-aware multi-object tracking, aiming to bridge the gap between vision and language by introducing unified challenges that evaluate both temporal localization and object tracking. To this end, we host a challenging competition where participants are required to develop models to tackle the SVAG task. The competition is hosted on the Codabench platform\footnote{\url{https://www.codabench.org/competitions/9743/}}, with SVAGEval as the official evaluation benchmark. Several teams have submitted their results, and a summary leaderboard is presented in Tab.~\ref{tab:competition-results}. We conducted several ablation studies and applied the best-performing spatial and temporal grounding results on the SVAGEval, and the final results are as SVAGFormer recorded in the table. We adopt \textbf{m-HIoU} as the primary ranking metric, defined as the arithmetic mean of HOTA and mIoU, which jointly capture temporal and spatial localization quality.

\begin{table}[!ht]
	\centering
	\renewcommand{\arraystretch}{1.1}
	\scalebox{0.84}{
		\begin{tabular}{lcccccc}
			\specialrule{1.5pt}{0pt}{0pt}
			%\rowcolor{mygray} 
			Participant & \textbf{m-HIoU} & HOTA & mIoU & DetA & AssA & R1@0.3 \\ \hline\hline
			xxcbole & \textbf{25.417} & 7.957 & \textbf{42.877} & 3.167 & 21.884 & \textbf{47.223} \\
y\_squared & 20.680 & \textbf{10.734} & 30.627 & 4.070 & \textbf{41.693} & 33.653 \\
gl0ria & 16.114 & 9.001 & 23.227 & 4.048 & 24.673 & 29.867 \\
SVAGFormer & 14.148 & 9.159 & 19.137 & \textbf{4.092} & 27.698 & 24.567 \\
\specialrule{1.5pt}{0pt}{0pt}
	       \end{tabular}}
	\caption{Competition leaderboard results on Codabench. Teams are ranked in descending order of \textbf{m-HIoU}. The highest score is highlighted in bold. The first team improved the overall performance by improving the temporal grounding performance. The second team improved the overall performance by improving the association.}
	\label{tab:competition-results}
	\vspace{-3mm}
\end{table}

Two teams (Team 1 and Team 3) added additional strategies and techniques to the baseline model to improve performance, and one team (Team 2) used additional models for tracking and retrieval to improve performance. These submissions reflect the growing interest in language-guided video understanding and reveal promising directions for improving spatio-temporal grounding, multimodal alignment, and long-horizon reasoning. We expect the SVAG challenge to serve as a catalyst for future research on scalable, action-aware, and temporally grounded vision-language models.
\vspace{-3mm}
% \begin{table}[h]
% 	\centering
% 	\renewcommand{\arraystretch}{1.1}
% 	\scalebox{0.44}{
% 		\begin{tabular}{rccccccccccccccccccc}
% 			\specialrule{1.5pt}{0pt}{0pt}
% 			\rowcolor{mygray} 
% 			Participant & \textbf{m-HIoU} & HOTA & mIoU & DetA & AssA & DetRe & DetPr & AssRe & AssPr & LocA & R1@0.1 & R1@0.3 & R1@0.5 & R5@0.1 & R5@0.3 & R5@0.5 & R10@0.1 & R10@0.3 & R10@0.5 \\ \hline\hline
% 			xxcbole & \textbf{25.417} & 7.957 & 42.877 & 3.167 & 21.884 & 4.794 & 7.332 & 30.178 & 28.604 & 68.989 & 66.203 & 47.223 & 41.437 & 66.203 & 47.223 & 41.437 & 66.203 & 47.223 & 41.437 \\
% y\_squared & 20.680 & 10.734 & 30.627 & 4.070 & 41.693 & 7.862 & 7.294 & 71.329 & 51.298 & 84.032 & 71.140 & 33.653 & 22.890 & 71.140 & 33.653 & 22.890 & 71.140 & 33.653 & 22.890 \\
% gl0ria & 16.114 & 9.001 & 23.227 & 4.048 & 24.673 & 6.840 & 7.939 & 34.225 & 38.885 & 71.152 & 47.620 & 29.867 & 17.467 & 57.880 & 36.680 & 25.157 & 76.653 & 49.457 & 30.620 \\
% our & 14.148 & 9.159 & 19.137 & 4.092 & 27.698 & 7.122 & 7.580 & 38.454 & 41.448 & 73.020 & 38.590 & 24.567 & 17.750 & 64.790 & 40.570 & 26.233 & 71.843 & 44.540 & 30.433 \\
% \specialrule{1.5pt}{0pt}{0pt}
% 	       \end{tabular}}
% 	\caption{Competition leaderboard results on Codabench. Teams are ranked in descending order of \textbf{m-HIoU}. The highest score is highlighted in bold \tanveer{add some insights from the figure, zoom in the table, remove unimportant metrics, don't need to add so many columns}}
% 	\label{tab:competition-results}
% 	\vspace{-3mm}
% \end{table}

\bibliographystyle{plain} 
\bibliography{references}

@article{lee2024enhancing,
  title={Enhancing Temporal Action Localization: Advanced s6 Modeling with Recurrent Mechanism},
  author={Lee, Sangyoun and Jung, Juho and Oh, Changdae and Yun, Sunghee},
  journal={arXiv preprint arXiv:2407.13078},
  year={2024}
}

@inproceedings{liu2024end,
  title={End-to-end temporal action detection with 1b parameters across 1000 frames},
  author={Liu, Shuming and Zhang, Chen-Lin and Zhao, Chen and Ghanem, Bernard},
  booktitle={Proceedings of the IEEE/CVF conference on computer vision and pattern recognition},
  pages={18591--18601},
  year={2024}
}

@article{shi2023temporal,
  title={Temporal action localization with enhanced instant discriminability},
  author={Shi, Dingfeng and Cao, Qiong and Zhong, Yujie and An, Shan and Cheng, Jian and Zhu, Haogang and Tao, Dacheng},
  journal={arXiv preprint arXiv:2309.05590},
  year={2023}
}

@inproceedings{khoreva2019video,
  title={Video object segmentation with language referring expressions},
  author={Khoreva, Anna and Rohrbach, Anna and Schiele, Bernt},
  booktitle={Computer Vision--ACCV 2018: 14th Asian Conference on Computer Vision, Perth, Australia, December 2--6, 2018, Revised Selected Papers, Part IV 14},
  pages={123--141},
  year={2019},
  organization={Springer}
}

@inproceedings{nagaraja2016modeling,
  title={Modeling context between objects for referring expression understanding},
  author={Nagaraja, Varun K and Morariu, Vlad I and Davis, Larry S},
  booktitle={Computer Vision--ECCV 2016: 14th European Conference, Amsterdam, The Netherlands, October 11--14, 2016, Proceedings, Part IV 14},
  pages={792--807},
  year={2016},
  organization={Springer}
}

@inproceedings{gu2024context,
  title={Context-guided spatio-temporal video grounding},
  author={Gu, Xin and Fan, Heng and Huang, Yan and Luo, Tiejian and Zhang, Libo},
  booktitle={Proceedings of the IEEE/CVF Conference on Computer Vision and Pattern Recognition},
  pages={18330--18339},
  year={2024}
}

@article{nguyen2023type,
  title={Type-to-track: Retrieve any object via prompt-based tracking},
  author={Nguyen, Pha and Quach, Kha Gia and Kitani, Kris and Luu, Khoa},
  journal={Advances in Neural Information Processing Systems},
  volume={36},
  pages={3205--3219},
  year={2023}
}

@inproceedings{fan2019lasot,
  title={Lasot: A high-quality benchmark for large-scale single object tracking},
  author={Fan, Heng and Lin, Liting and Yang, Fan and Chu, Peng and Deng, Ge and Yu, Sijia and Bai, Hexin and Xu, Yong and Liao, Chunyuan and Ling, Haibin},
  booktitle={Proceedings of the IEEE/CVF conference on computer vision and pattern recognition},
  pages={5374--5383},
  year={2019}
}

@misc{openai2023,
  author       = {{OpenAI}},
  year         = {2023},
  howpublished = {\url{https://chatgpt.com/}},
}

@article{dendorfer2021motchallenge,
  title={Motchallenge: A benchmark for single-camera multiple target tracking},
  author={Dendorfer, Patrick and Osep, Aljosa and Milan, Anton and Schindler, Konrad and Cremers, Daniel and Reid, Ian and Roth, Stefan and Leal-Taix{\'e}, Laura},
  journal={International Journal of Computer Vision},
  volume={129},
  pages={845--881},
  year={2021},
  publisher={Springer}
}

@article{dendorfer2020mot20,
  title={Mot20: A benchmark for multi object tracking in crowded scenes},
  author={Dendorfer, Patrick and Rezatofighi, Hamid and Milan, Anton and Shi, Javen and Cremers, Daniel and Reid, Ian and Roth, Stefan and Schindler, Konrad and Leal-Taix{\'e}, Laura},
  journal={arXiv preprint arXiv:2003.09003},
  year={2020}
}

@article{qi2022occluded,
  title={Occluded video instance segmentation: A benchmark},
  author={Qi, Jiyang and Gao, Yan and Hu, Yao and Wang, Xinggang and Liu, Xiaoyu and Bai, Xiang and Belongie, Serge and Yuille, Alan and Torr, Philip HS and Bai, Song},
  journal={International Journal of Computer Vision},
  volume={130},
  number={8},
  pages={2022--2039},
  year={2022},
  publisher={Springer}
}

@inproceedings{seo2020urvos,
  title={Urvos: Unified referring video object segmentation network with a large-scale benchmark},
  author={Seo, Seonguk and Lee, Joon-Young and Han, Bohyung},
  booktitle={Computer Vision--ECCV 2020: 16th European Conference, Glasgow, UK, August 23--28, 2020, Proceedings, Part XV 16},
  pages={208--223},
  year={2020},
  organization={Springer}
}

@inproceedings{dave2020tao,
  title={Tao: A large-scale benchmark for tracking any object},
  author={Dave, Achal and Khurana, Tarasha and Tokmakov, Pavel and Schmid, Cordelia and Ramanan, Deva},
  booktitle={European conference on computer vision},
  pages={436--454},
  year={2020},
  organization={Springer}
}

@article{lei2019tvqa+,
  title={Tvqa+: Spatio-temporal grounding for video question answering},
  author={Lei, Jie and Yu, Licheng and Berg, Tamara L and Bansal, Mohit},
  journal={arXiv preprint arXiv:1904.11574},
  year={2019}
}

@article{xu2022point,
  title={Point-supervised video temporal grounding},
  author={Xu, Zhe and Wei, Kun and Yang, Xu and Deng, Cheng},
  journal={IEEE Transactions on Multimedia},
  volume={25},
  pages={6121--6131},
  year={2022},
  publisher={IEEE}
}

@article{chen2021end,
  title={End-to-end multi-modal video temporal grounding},
  author={Chen, Yi-Wen and Tsai, Yi-Hsuan and Yang, Ming-Hsuan},
  journal={Advances in Neural Information Processing Systems},
  volume={34},
  pages={28442--28453},
  year={2021}
}

@inproceedings{lin2023univtg,
  title={Univtg: Towards unified video-language temporal grounding},
  author={Lin, Kevin Qinghong and Zhang, Pengchuan and Chen, Joya and Pramanick, Shraman and Gao, Difei and Wang, Alex Jinpeng and Yan, Rui and Shou, Mike Zheng},
  booktitle={Proceedings of the IEEE/CVF International Conference on Computer Vision},
  pages={2794--2804},
  year={2023}
}

@article{luo2025spatial,
  title={Spatial--temporal video grounding with cross-modal understanding and enhancement},
  author={Luo, Shu and Pan, Jingyu and Cao, Da and Wang, Jiawei and Le, Yuquan and Liu, Meng},
  journal={Expert Systems with Applications},
  volume={271},
  pages={126650},
  year={2025},
  publisher={Elsevier}
}

@article{wang2025spacevllm,
  title={SpaceVLLM: Endowing Multimodal Large Language Model with Spatio-Temporal Video Grounding Capability},
  author={Wang, Jiankang and Zhang, Zhihan and Liu, Zhihang and Li, Yang and Ge, Jiannan and Xie, Hongtao and Zhang, Yongdong},
  journal={arXiv preprint arXiv:2503.13983},
  year={2025}
}

@inproceedings{lin2022stvgformer,
  title={Stvgformer: Spatio-temporal video grounding with static-dynamic cross-modal understanding},
  author={Lin, Zihang and Tan, Chaolei and Hu, Jian-Fang and Jin, Zhi and Ye, Tiancai and Zheng, Wei-Shi},
  booktitle={Proceedings of the 4th on Person in Context Workshop},
  pages={1--5},
  year={2022}
}

@inproceedings{xiao2020visual,
  title={Visual relation grounding in videos},
  author={Xiao, Junbin and Shang, Xindi and Yang, Xun and Tang, Sheng and Chua, Tat-Seng},
  booktitle={European conference on computer vision},
  pages={447--464},
  year={2020},
  organization={Springer}
}

@inproceedings{zhang2020does,
  title={Where does it exist: Spatio-temporal video grounding for multi-form sentences},
  author={Zhang, Zhu and Zhao, Zhou and Zhao, Yang and Wang, Qi and Liu, Huasheng and Gao, Lianli},
  booktitle={Proceedings of the IEEE/CVF Conference on Computer Vision and Pattern Recognition},
  pages={10668--10677},
  year={2020}
}

@article{tang2021human,
  title={Human-centric spatio-temporal video grounding with visual transformers},
  author={Tang, Zongheng and Liao, Yue and Liu, Si and Li, Guanbin and Jin, Xiaojie and Jiang, Hongxu and Yu, Qian and Xu, Dong},
  journal={IEEE Transactions on Circuits and Systems for Video Technology},
  volume={32},
  number={12},
  pages={8238--8249},
  year={2021},
  publisher={IEEE}
}

@inproceedings{shang2019annotating,
  title={Annotating objects and relations in user-generated videos},
  author={Shang, Xindi and Di, Donglin and Xiao, Junbin and Cao, Yu and Yang, Xun and Chua, Tat-Seng},
  booktitle={Proceedings of the 2019 on International Conference on Multimedia Retrieval},
  pages={279--287},
  year={2019}
}

@article{lei2021detecting,
  title={Detecting moments and highlights in videos via natural language queries},
  author={Lei, Jie and Berg, Tamara L and Bansal, Mohit},
  journal={Advances in Neural Information Processing Systems},
  volume={34},
  pages={11846--11858},
  year={2021}
}

@inproceedings{gao2017tall,
  title={Tall: Temporal activity localization via language query},
  author={Gao, Jiyang and Sun, Chen and Yang, Zhenheng and Nevatia, Ram},
  booktitle={Proceedings of the IEEE international conference on computer vision},
  pages={5267--5275},
  year={2017}
}

@article{regneri2013grounding,
  title={Grounding action descriptions in videos},
  author={Regneri, Michaela and Rohrbach, Marcus and Wetzel, Dominikus and Thater, Stefan and Schiele, Bernt and Pinkal, Manfred},
  journal={Transactions of the Association for Computational Linguistics},
  volume={1},
  pages={25--36},
  year={2013},
  publisher={MIT Press One Rogers Street, Cambridge, MA 02142-1209, USA journals-info~…}
}

@inproceedings{wu2023referring,
  title={Referring multi-object tracking},
  author={Wu, Dongming and Han, Wencheng and Wang, Tiancai and Dong, Xingping and Zhang, Xiangyu and Shen, Jianbing},
  booktitle={Proceedings of the IEEE/CVF conference on computer vision and pattern recognition},
  pages={14633--14642},
  year={2023}
}

@article{zhang2024bootstrapping,
  title={Bootstrapping Referring Multi-Object Tracking},
  author={Zhang, Yani and Wu, Dongming and Han, Wencheng and Dong, Xingping},
  journal={arXiv preprint arXiv:2406.05039},
  year={2024}
}

@article{zhang2023dvisplus,
  title={DVIS++: Improved Decoupled Framework for Universal Video Segmentation}, 
  author={Tao Zhang and Xingye Tian and Yikang Zhou and Shunping Ji and Xuebo Wang and Xin Tao and Yuan Zhang and Pengfei Wan and Zhongyuan Wang and Yu Wu},
  journal={arXiv preprint arXiv:2312.13305},
  year={2023},
}

@InProceedings{Cao_2025_WACV,
    author    = {Cao, Zhuo and Zhang, Bingqing and Du, Heming and Yu, Xin and Li, Xue and Wang, Sen},
    title     = {FlashVTG: Feature Layering and Adaptive Score Handling Network for Video Temporal Grounding},
    booktitle = {Proceedings of the Winter Conference on Applications of Computer Vision (WACV)},
    month     = {February},
    year      = {2025},
    pages     = {9208-9218}
}

@inproceedings{qu2024chatvtg,
  title={Chatvtg: Video temporal grounding via chat with video dialogue large language models},
  author={Qu, Mengxue and Chen, Xiaodong and Liu, Wu and Li, Alicia and Zhao, Yao},
  booktitle={Proceedings of the IEEE/CVF Conference on Computer Vision and Pattern Recognition},
  pages={1847--1856},
  year={2024}
}

@article{luiten2021hota,
  title={Hota: A higher order metric for evaluating multi-object tracking},
  author={Luiten, Jonathon and Osep, Aljosa and Dendorfer, Patrick and Torr, Philip and Geiger, Andreas and Leal-Taix{\'e}, Laura and Leibe, Bastian},
  journal={International journal of computer vision},
  volume={129},
  pages={548--578},
  year={2021},
  publisher={Springer}
}

@inproceedings{jiang2024prior,
  title={Prior knowledge integration via llm encoding and pseudo event regulation for video moment retrieval},
  author={Jiang, Yiyang and Zhang, Wengyu and Zhang, Xulu and Wei, Xiao-Yong and Chen, Chang Wen and Li, Qing},
  booktitle={Proceedings of the 32nd ACM International Conference on Multimedia},
  pages={7249--7258},
  year={2024}
}

@inproceedings{wang2024internvideo2,
  title={Internvideo2: Scaling foundation models for multimodal video understanding},
  author={Wang, Yi and Li, Kunchang and Li, Xinhao and Yu, Jiashuo and He, Yinan and Chen, Guo and Pei, Baoqi and Zheng, Rongkun and Wang, Zun and Shi, Yansong and others},
  booktitle={European Conference on Computer Vision},
  pages={396--416},
  year={2024},
  organization={Springer}
}

@article{touvron2023llama,
  title={Llama 2: Open foundation and fine-tuned chat models},
  author={Touvron, Hugo and Martin, Louis and Stone, Kevin and Albert, Peter and Almahairi, Amjad and Babaei, Yasmine and Bashlykov, Nikolay and Batra, Soumya and Bhargava, Prajjwal and Bhosale, Shruti and others},
  journal={arXiv preprint arXiv:2307.09288},
  year={2023}
}

@inproceedings{liu2024r,
  title={R 2-Tuning: Efficient Image-to-Video Transfer Learning for Video Temporal Grounding},
  author={Liu, Ye and He, Jixuan and Li, Wanhua and Kim, Junsik and Wei, Donglai and Pfister, Hanspeter and Chen, Chang Wen},
  booktitle={European Conference on Computer Vision},
  pages={421--438},
  year={2024},
  organization={Springer}
}

@article{zhao2025ld,
  title={Ld-detr: Loop decoder detection transformer for video moment retrieval and highlight detection},
  author={Zhao, Pengcheng and He, Zhixian and Zhang, Fuwei and Lin, Shujin and Zhou, Fan},
  journal={arXiv preprint arXiv:2501.10787},
  year={2025}
}

@inproceedings{li2025language,
  title={Language Decoupling with Fine-grained Knowledge Guidance for Referring Multi-object Tracking},
  author={Li, Guangyao and Zhuang, Siping and Jian, Yajun and Yan, Yan and Wang, Hanzi},
  booktitle={Proceedings of the IEEE/CVF International Conference on Computer Vision},
  pages={23626--23635},
  year={2025}
}

@misc{qwen2.5-VL,
    title = {Qwen2.5-VL},
    url = {https://qwenlm.github.io/blog/qwen2.5-vl/},
    author = {Qwen Team},
    month = {January},
    year = {2025}
}

@misc{qwen3technicalreport,
      title={Qwen3 Technical Report}, 
      author={Qwen Team},
      year={2025},
      eprint={2505.09388},
      archivePrefix={arXiv},
      primaryClass={cs.CL},
      url={https://arxiv.org/abs/2505.09388}, 
}

@inproceedings{huang2024vtimellm,
  title={Vtimellm: Empower llm to grasp video moments},
  author={Huang, Bin and Wang, Xin and Chen, Hong and Song, Zihan and Zhu, Wenwu},
  booktitle={Proceedings of the IEEE/CVF Conference on Computer Vision and Pattern Recognition},
  pages={14271--14280},
  year={2024}
}

@article{yuan2025sa2va,
  title={Sa2va: Marrying sam2 with llava for dense grounded understanding of images and videos},
  author={Yuan, Haobo and Li, Xiangtai and Zhang, Tao and Sun, Yueyi and Huang, Zilong and Xu, Shilin and Ji, Shunping and Tong, Yunhai and Qi, Lu and Feng, Jiashi and others},
  journal={arXiv preprint arXiv:2501.04001},
  year={2025}
}

@inproceedings{zeng2025factorized,
  title={Factorized Learning for Temporally Grounded Video-Language Models},
  author={Zeng, Wenzheng and Gao, Difei and Shou, Mike Zheng and Ng, Hwee Tou},
  booktitle={Proceedings of the IEEE/CVF International Conference on Computer Vision},
  pages={20683--20693},
  year={2025}
}

@article{zeng2024timesuite,
  title={Timesuite: Improving mllms for long video understanding via grounded tuning},
  author={Zeng, Xiangyu and Li, Kunchang and Wang, Chenting and Li, Xinhao and Jiang, Tianxiang and Yan, Ziang and Li, Songze and Shi, Yansong and Yue, Zhengrong and Wang, Yi and others},
  journal={arXiv preprint arXiv:2410.19702},
  year={2024}
}

@article{liang2025finegrained,
      title={Fine-grained Spatiotemporal Grounding on Egocentric Videos}, 
      author={Shuo Liang and Yiwu Zhong and Zi-Yuan Hu and Yeyao Tao and Liwei Wang},
      journal={arxiv preprint arXiv:2508.00518},
      year={2025},
}

@article{bai2024one,
  title={One token to seg them all: Language instructed reasoning segmentation in videos},
  author={Bai, Zechen and He, Tong and Mei, Haiyang and Wang, Pichao and Gao, Ziteng and Chen, Joya and Liu, Lei and Zhang, Zheng and Shou, Mike Zheng},
  journal={arXiv preprint arXiv:2409.19603},
  year={2024}
}

@article{wang2025internvl3_5,
  title={InternVL3.5: Advancing Open-Source Multimodal Models in Versatility, Reasoning, and Efficiency},
  author={Wang, Weiyun and Gao, Zhangwei and Gu, Lixin and Pu, Hengjun and Cui, Long and Wei, Xingguang and Liu, Zhaoyang and Jing, Linglin and Ye, Shenglong and Shao, Jie and others},
  journal={arXiv preprint arXiv:2508.18265},
  year={2025}
}

@misc{vteam2025glm45vglm41vthinkingversatilemultimodal,
      title={GLM-4.5V and GLM-4.1V-Thinking: Towards Versatile Multimodal Reasoning with Scalable Reinforcement Learning}, 
      author={V Team and Wenyi Hong and Wenmeng Yu and Xiaotao Gu and Guo Wang and Guobing Gan and Haomiao Tang and Jiale Cheng and Ji Qi and Junhui Ji and Lihang Pan and Shuaiqi Duan and Weihan Wang and Yan Wang and Yean Cheng and Zehai He and Zhe Su and Zhen Yang and Ziyang Pan and Aohan Zeng and Baoxu Wang and Bin Chen and Boyan Shi and Changyu Pang and Chenhui Zhang and Da Yin and Fan Yang and Guoqing Chen and Jiazheng Xu and Jiale Zhu and Jiali Chen and Jing Chen and Jinhao Chen and Jinghao Lin and Jinjiang Wang and Junjie Chen and Leqi Lei and Letian Gong and Leyi Pan and Mingdao Liu and Mingde Xu and Mingzhi Zhang and Qinkai Zheng and Sheng Yang and Shi Zhong and Shiyu Huang and Shuyuan Zhao and Siyan Xue and Shangqin Tu and Shengbiao Meng and Tianshu Zhang and Tianwei Luo and Tianxiang Hao and Tianyu Tong and Wenkai Li and Wei Jia and Xiao Liu and Xiaohan Zhang and Xin Lyu and Xinyue Fan and Xuancheng Huang and Yanling Wang and Yadong Xue and Yanfeng Wang and Yanzi Wang and Yifan An and Yifan Du and Yiming Shi and Yiheng Huang and Yilin Niu and Yuan Wang and Yuanchang Yue and Yuchen Li and Yutao Zhang and Yuting Wang and Yu Wang and Yuxuan Zhang and Zhao Xue and Zhenyu Hou and Zhengxiao Du and Zihan Wang and Peng Zhang and Debing Liu and Bin Xu and Juanzi Li and Minlie Huang and Yuxiao Dong and Jie Tang},
      year={2025},
      eprint={2507.01006},
      archivePrefix={arXiv},
      primaryClass={cs.CV},
      url={https://arxiv.org/abs/2507.01006}, 
}

@inproceedings{hannan2024rgnet,
  title={Rgnet: A unified clip retrieval and grounding network for long videos},
  author={Hannan, Tanveer and Islam, Md Mohaiminul and Seidl, Thomas and Bertasius, Gedas},
  booktitle={European Conference on Computer Vision},
  pages={352--369},
  year={2024},
  organization={Springer}
}

@inproceedings{hannan2025revisionllm,
  title={Revisionllm: Recursive vision-language model for temporal grounding in hour-long videos},
  author={Hannan, Tanveer and Islam, Md Mohaiminul and Gu, Jindong and Seidl, Thomas and Bertasius, Gedas},
  booktitle={Proceedings of the IEEE/CVF Conference on Computer Vision and Pattern Recognition},
  pages={19012--19022},
  year={2025}
}

@article{hannan2023gratt,
  title={Gratt-vis: Gated residual attention for auto rectifying video instance segmentation},
  author={Hannan, Tanveer and Koner, Rajat and Bernhard, Maximilian and Shit, Suprosanna and Menze, Bjoern and Tresp, Volker and Schubert, Matthias and Seidl, Thomas},
  journal={arXiv preprint arXiv:2305.17096},
  year={2023}
}

@inproceedings{koner2023instanceformer,
  title={Instanceformer: An online video instance segmentation framework},
  author={Koner, Rajat and Hannan, Tanveer and Shit, Suprosanna and Sharifzadeh, Sahand and Schubert, Matthias and Seidl, Thomas and Tresp, Volker},
  booktitle={Proceedings of the AAAI Conference on Artificial Intelligence},
  volume={37},
  number={1},
  pages={1188--1195},
  year={2023}
}

@article{hannan2022box,
  title={Box supervised video segmentation proposal network},
  author={Hannan, Tanveer and Koner, Rajat and Kobold, Jonathan and Schubert, Matthias},
  journal={arXiv preprint arXiv:2202.07025},
  year={2022}
}

@article{hannan2025docslm,
  title={DocSLM: A Small Vision-Language Model for Long Multimodal Document Understanding},
  author={Hannan, Tanveer and Mallios, Dimitrios and Pathak, Parth and Sardari, Faegheh and Seidl, Thomas and Bertasius, Gedas and Fayyaz, Mohsen and Sengupta, Sunando},
  journal={arXiv preprint arXiv:2511.11313},
  year={2025}
}

@misc{openai_gpt41mini_2025,
  title        = {Introducing GPT-4.1},
  author       = {{OpenAI}},
  year         = {2025},
  howpublished = {\url{https://openai.com/index/gpt-4-1/}},
    note = "[Online; accessed 2026-05-06]"
}

@misc{openai_gpt54_2026,
  title        = {Introducing GPT-5.4},
  author       = {{OpenAI}},
  year         = {2026},
  howpublished = {\url{https://openai.com/index/introducing-gpt-5-4/}},
    note = "[Online; accessed 2026-05-06]"
}

@misc{anthropic2026claudeopus46,
  author       = {Anthropic},
  title        = {Introducing Claude Opus 4.6},
  year         = {2026},
  month        = feb,
  day          = {5},
  url          = {https://www.anthropic.com/news/claude-opus-4-6},
  note = "[Online; accessed 2026-05-06]"
}
\end{document}